%% file: main.tex
\documentclass{article}

\PassOptionsToPackage{numbers,compress}{natbib}
\usepackage[eandd,preprint]{neurips_2026}

\usepackage[utf8]{inputenc}
\usepackage[T1]{fontenc}
\usepackage{courier}
\usepackage{hyperref}
\usepackage{url}
\usepackage{booktabs}
\usepackage{microtype}
\usepackage[table]{xcolor}
\usepackage{tabularx}
\usepackage[normalem]{ulem}
\usepackage{array}
\usepackage{graphicx}
\usepackage{amsmath}
\usepackage{amssymb}
\usepackage{enumitem}
\usepackage{pifont}
\usepackage{float}
\usepackage{wrapfig}
\usepackage[capitalise,nameinlink]{cleveref}

\newcommand{\benchmark}{\textsc{CUDABeaver}}

\newcommand{\passone}{pass@1}
\newcommand{\passk}{pass@\ensuremath{k}}
\newcommand{\debugk}{debug\_rate@\ensuremath{k}}
\definecolor{goodgreen}{RGB}{34,139,34}
\definecolor{badred}{RGB}{180,38,38}
\newcommand{\cmark}{\textcolor{goodgreen}{\ding{51}}}
\newcommand{\xmark}{\textcolor{badred}{\ding{55}}}
\newcommand{\numtasks}{213}

\newcommand{\numcategories}{8}
\newcommand{\nummetrics}{4}

\title{CUDABeaver: Benchmarking LLM-Based Automated CUDA Debugging}
\author{%
  Shiyang Li\thanks{Equal contribution.} \\
  University of Minnesota \\
  \texttt{li004074@umn.edu} \\
  \And
  Haoyang Chen\footnotemark[1] \\
  University of Minnesota \\
  \texttt{chen9861@umn.edu} \\
  \AND
  Mattia Fazzini \\
  University of Minnesota \\
  \texttt{mfazzini@umn.edu} \\
  \And
  Caiwen Ding \\
  University of Minnesota \\
  \texttt{dingc@umn.edu}
}
\date{}

\begin{document}
\maketitle

\input{sections/abstract}
\input{sections/introduction}
\input{sections/related}
\input{sections/eval}        
\input{sections/experiments} 
\input{sections/conclusion}  
\newpage
\bibliographystyle{plainnat}
\bibliography{references}

\newpage
\input{sections/appendix}

\end{document}

%% file: sections/abstract.tex
\begin{abstract}
Debugging CUDA programs has long been challenging because failures often arise from subtle interactions among hardware behavior, compiler decisions, memory hierarchy, and asynchronous execution. More importantly, with the rapid expansion of GPU usage across scientific computing, machine learning, graphics, and systems workloads, CUDA debugging has become more challenging than ever. Current evaluations of LLM-based CUDA programming largely miss this setting: a model can pass correctness tests with \emph{repair by degeneration}, simplifying the CUDA code into a safer but slower program that abandons the original optimization structure. We introduce \benchmark{}, a benchmark for CUDA debugging from real failing workspaces produced during LLM-based CUDA generation. Each task provides the broken candidate, native build/test commands, raw error evidence, and a single editable file.
\benchmark{} evaluates whether a fixer truly repairs the failing CUDA code or merely finds a slower test-passing replacement, reporting results by failure category, debugging trajectory, stagnation mode, and performance preservation. We further propose $\text{pass@}k(M, C, A)$, a protocol-conditional CUDA debugging metric by making the fixer $M$, corpus $C$, and protocol axes $A$ explicit. Using this metric across \numtasks{} tasks and seven frontier LLMs, we show that protocol-aware evaluation gives a more faithful view of CUDA debugging ability: when performance-loss tolerance is high, fixers appear much stronger, but even a minor stricter performance requirement can sharply reduce measured success, shifting scores by up to 40 percentage points. Code and data are available at \url{https://github.com/HaoyangChen23/CUDABeaver}.
\end{abstract}

%% file: sections/introduction.tex
\section{Introduction}
\label{sec:intro}

Large language models (LLMs) have shown promising progress in automated GPU programming~\cite{avo,stitchcuda,cudaforge,liu2026dr}. Most current related benchmarks evaluate this progress as \emph{kernel generation}: a model generates a CUDA kernel from a task specification, scored by correctness and performance tests~\citep{ouyang2025kernelbench,zhu2026cudabench,computeeval2025,li2025tritonbench}. This framing measures generation ability but misses \emph{CUDA debugging}: repairing a concrete failing kernel without losing the optimization structure that makes it useful.

\begin{wrapfigure}{r}{0.6\linewidth}
    \vspace{-1.0em}
    \centering
    \includegraphics[width=\linewidth]{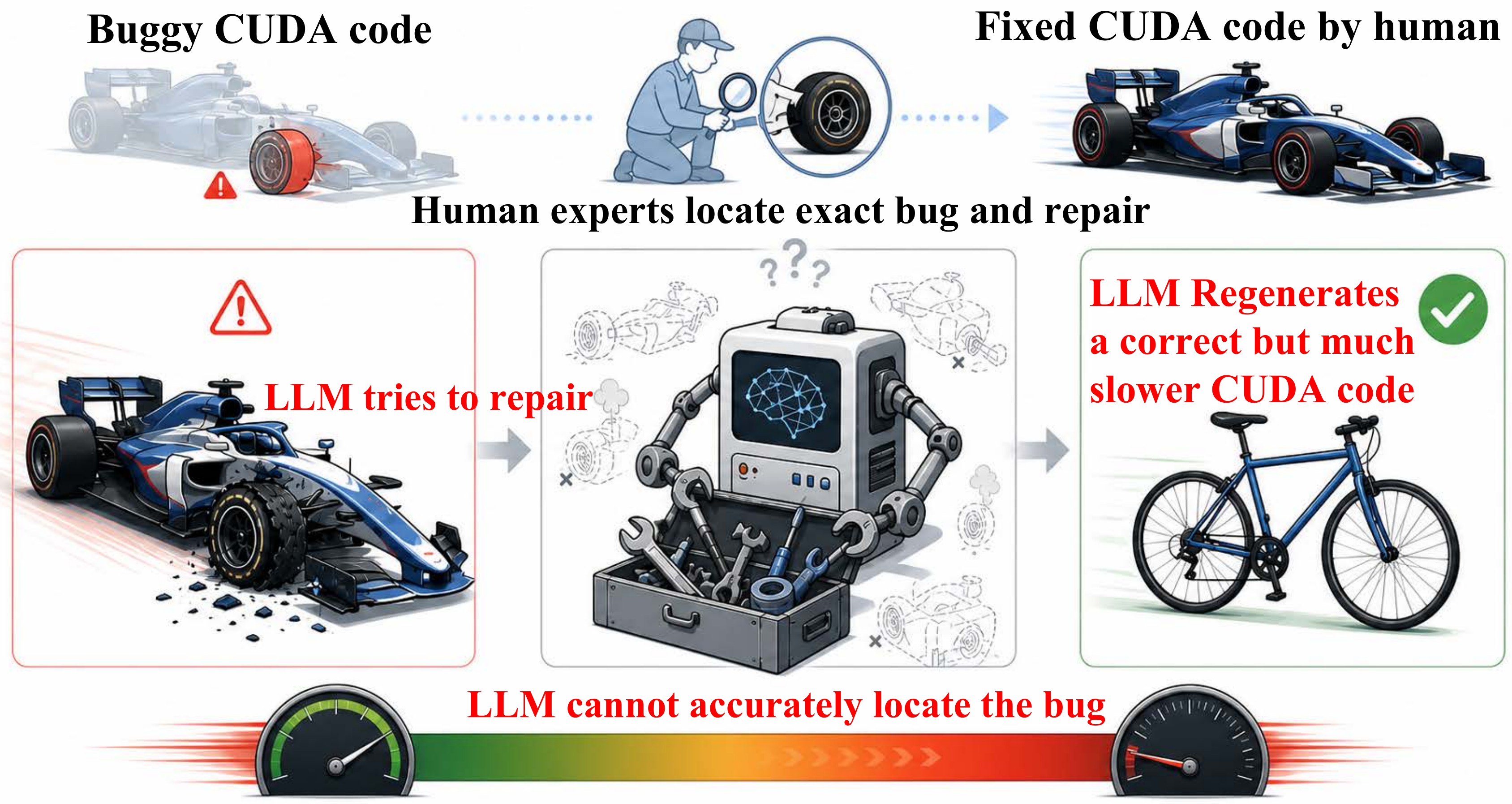}
    \vspace{-1.4em}
    \caption{\textbf{Repair by degeneration.} Here, the racecar denotes an optimized GPU kernel and the bicycle a correct-but-slow fallback; a useful repair preserves optimization structure, but LLMs often simplify candidates into slower correct programs.}
    \label{fig:repair-degeneration}
    \vspace{-1.2em}
\end{wrapfigure}

Debugging CUDA code is not ordinary program repair. Failures emerge from subtle interactions in a heterogeneous GPU execution model, including thread-level parallelism, explicit memory spaces, and asynchronous execution with host-device synchronization, and their signals are often non-local: a bounds error may surface only at a later CUDA API call. Prior CUDA debugging research targets memory-safety violations, data races, and library-level bugs with specialized detectors~\citep{tarek2023cucatch,tarek2026hunting,zhou2025fuzz4cuda}. A useful repair, however, must also preserve performance: it cannot serialize execution, over-synchronize, or replace an optimized kernel with a slower fallback. We therefore cast CUDA debugging as a \emph{hardware-aware, performance-preserving repair} problem, distinct from both general-purpose program repair and CUDA code generation. We call the resulting CUDA-specific LLM failure mode \emph{repair by degeneration}: the model satisfies the test harness by abandoning the original kernel's optimized structure rather than diagnosing the concrete failure (\Cref{fig:repair-degeneration}), yielding patches that pass narrow tests while degrading performance in ways aggregate \passk{} cannot expose.

Existing benchmarks cannot expose \emph{repair by degeneration} behavior of LLMs. On these benchmarks, evaluation starts from specifications and discards the failing intermediate programs that would reveal debug behavior, conflating generation with later repair. General code debugging benchmarks start from buggy code but do not cover CUDA programs~\citep{tian2024debugbench,jimenez2023swe}. Further, the widely used metric \passk{} on existing benchmarks also cannot evaluate and expose this. 

To fill this gap, we introduce \benchmark{}: a corpus of \numtasks{} reproducible CUDA debugging cases, each starts from a real failing workspace and labeled with one of eight evidence-grounded failure categories. On this corpus, repair can be evaluated beyond aggregate \passk{}: by failure category, by iteration-to-fix, and by whether the repair preserves the original kernel's performance intent. 

We formalise that \passk{} itself is protocol-conditional. Existing benchmarks report a single leaderboard while silently ignoring \emph{repair by degeneration}, test-time sampling, feedback richness, and conversation-history depth; we treat these as first-class experimental variables and propose $\text{pass@}k(M, C, A)$, making model $M$, corpus $C$, and protocol axes $A$ explicit, where $A$ is a multi-dimensional evaluation-axis vector (\S\ref{sec:benchmark:sensitivity}). To complement score-level sensitivity, we also report Kendall's $\tau$ to measure how stable cross-model rankings remain under different protocol settings.

Evaluation of seven frontier LLMs on \benchmark{} yields single-axis swings up to 40.0 percentage points and Kendall $\tau \in [0.47, 0.73]$ in cross-model rankings; notably, the smallest-swing axis (sampling method, $\Delta = 9.9$ percentage points) yields the strongest reshuffle ($\tau = 0.47$), indicating the score movement and ranking stability can decouple even when headline deltas appear modest.

This paper makes five contributions: (1) we formulate \emph{CUDA debugging} as a hardware-aware, performance-preserving repair problem that is distinct from both CUDA code generation and general-purpose program repair; (2) we identify \emph{repair by degeneration} as a CUDA-specific LLM failure mode, in which a model obtains a safer program by abandoning the optimized structure of the original code without diagnosing the concrete failure; (3) we release \benchmark{}, a corpus of \numtasks{} reproducible debugging cases, each exposing a constrained edit surface, native reproduction commands, and raw diagnostic evidence; (4) we introduce an evidence-grounded eight-category CUDA failure taxonomy and evaluation protocol for measuring category-specific repair, debugging trajectories, and performance-preserving fixes rather than only aggregate \passk{}; and (5) we formalise protocol-conditional $\text{pass@}k(M, C, A)$ along four evaluation axes, report per-axis swing $\Delta$ and cross-model Kendall $\tau$ as concrete robustness metrics, and propose an evaluation card schema for disclosing the protocol stack behind each reported \passk{}.

%% file: sections/related.tex
\section{Related Work}
\label{sec:related}

\paragraph{General code debugging and program repair.}
LLM-based code debugging is well studied in general-purpose programming. DebugBench~\citep{tian2024debugbench} evaluates repair on buggy C++, Java, and Python programs; SWE-bench~\citep{jimenez2023swe} asks models to resolve real GitHub issues; Self-Debugging~\citep{madaan2023selfrefineiterativerefinementselffeedback} studies iterative repair with execution feedback; and ChatRepair-style systems~\citep{olausson2023self} use runtime evidence to improve fixes. These benchmarks show that feedback can aid repair, but they center on CPU-side.

\paragraph{CUDA-specific bug detection and debugging research.}
CUDA programs exhibit persistent failure modes that motivate specialized analyses, including memory-safety violations, races, and vulnerabilities in GPU libraries \citep{wu2019characterizing,tarek2023cucatch,tarek2026hunting,zhou2025fuzz4cuda}. Wu et al.\ provide a foundational taxonomy of 319 real CUDA bugs from GitHub \citep{wu2019characterizing}. Subsequent systems leverage large-scale fuzzing for CUDA debugging \citep{tarek2023cucatch,tarek2026hunting,zhou2025fuzz4cuda}. None of existing works discuss LLM-based CUDA debugging.

\paragraph{GPU program generation benchmarks.}
The closest CUDA-oriented benchmarks (KernelBench~\citep{ouyang2025kernelbench}, Compute-Eval~\citep{computeeval2025}, and CUDABench~\citep{zhu2026cudabench}) primarily study specification-to-kernel generation from task descriptions or reference PyTorch API. These benchmarks measure generation quality of LLMs, but ignore the debugging ability.

\paragraph{LLM-based CUDA generation systems.}
Recent agentic CUDA systems use compile, execution, and profiling feedback to improve generated kernels~\citep{cudaforge,stitchcuda,avo}. They show that iterative feedback is valuable for LLM-based GPU programming. StitchCUDA~\cite{stitchcuda} and Dr. Kernel~\cite{liu2026dr} also mention the degeneration behavior of LLMs but do not quantify how it impacts CUDA generation quality, because no evaluation metric for this is practical on benchmarks they used.


\paragraph{Protocol-conditional evaluation in LLM benchmarking.} Recent work shows that benchmark scores often reflect evaluation protocol as much as model capability. \citet{sclar2023quantifying} report a 76 percentage point shift for LLaMA-2-13B under prompt-format changes, \citet{alzahrani2024benchmarks} show that MMLU rankings can move by up to 8 positions as protocol choices vary, and \citet{mizrahi2024state} argue that a prompt outcome should be viewed as a single draw from a broader response distribution. \benchmark{} applies this lens to CUDA debugging through $\text{pass@}k(M, C, A)$, quantifying dependence on four protocol axes and reporting per-axis swing $\Delta$ together with Kendall $\tau$ as robustness measures (\S\ref{sec:evaluation}).

\paragraph{Positioning.} Relative to prior work, \benchmark{} occupies a distinct point in the design space by combining broken-start tasks, CUDA-native workloads, repair-focused evaluation, and explicit protocol conditionality. Prior work variously studies general code repair, CUDA bug detection, or CUDA generation; none combine a fixed-broken-start protocol with repair-focused evaluation. Protocol-sensitivity studies motivate this framing but have not instantiated it as a CUDA debugging benchmark. \benchmark{} fills that gap for measuring whether models can produce performance-preserving repairs without degenerating into a slower fallback.

%% file: sections/eval.tex
\section{Benchmark Design}
\label{sec:benchmark}

\benchmark{} makes the CUDA debugging skill, hidden by single-number \passk{}, explicit and testable: debugging is decoupled from generation via the broken-start protocol over \numtasks{} curated and generation-benchmark tasks; $\text{pass@}k(M, C, A)$ and Kendall $\tau$ then characterise per-model capability and cross-model ranking robustness across protocol axes.

\subsection{Debug-from-Broken-Start Protocol}
\label{sec:benchmark:protocol}

\paragraph{Why \emph{generate-then-iterate} fuses two skills.}
A standard iterative protocol asks the model to write a CUDA code from scratch on iteration~1 and, on subsequent iterations, to improve its own previous output. During this process, regeneration or debugging is necessary when model generates a failed code. The headline metric \passk{} therefore mixes two distinct capabilities: producing a reasonable initial CUDA kernel from a specification (\emph{generation}), and recovering from a concrete failing kernel under a concrete error signal (\emph{debugging}).

\begin{figure}[t]
    \centering
    \includegraphics[width=\linewidth]{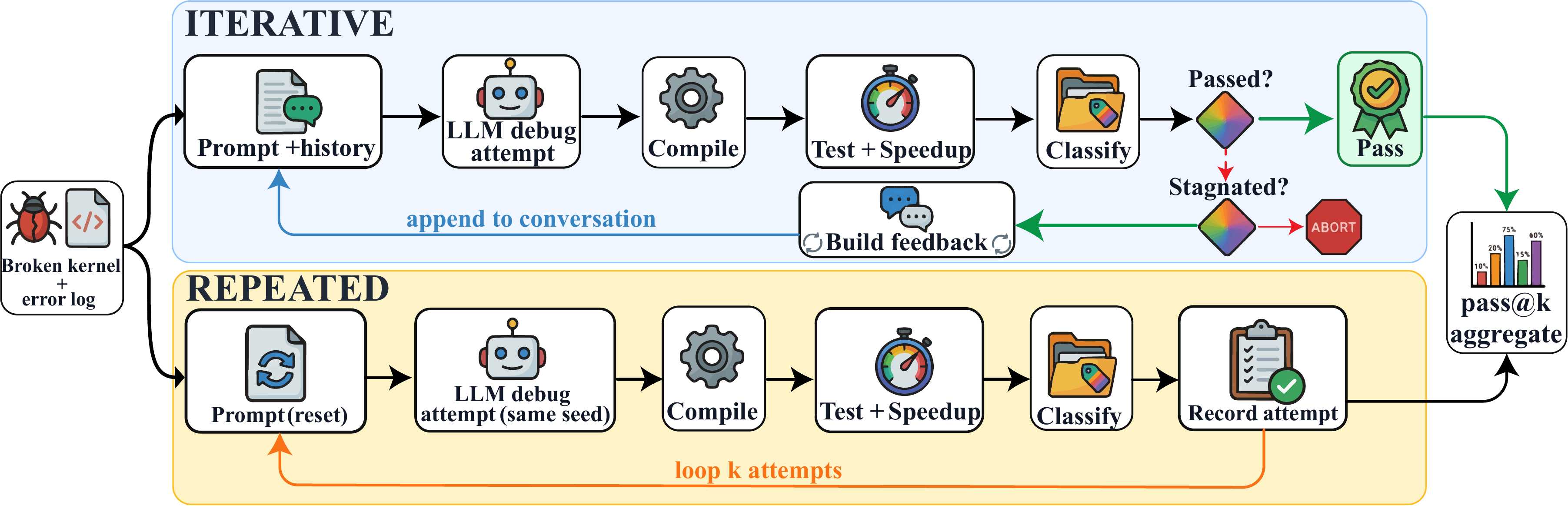}
    \caption{Iterative vs.\ repeated debugging pipelines (axis $A_2$). The iterative pipeline appends feedback after each failed attempt and exits early on success; the repeated pipeline resets to the initial prompt each attempt. Both feed identical outputs to the same \passk{} aggregator.}
    \label{fig:pipeline-iter-vs-repeated}
\end{figure}

\paragraph{The debug-from-broken-start protocol.}
\benchmark{} re-frames the task by removing the generation step entirely.
Every task in the corpus carries a \emph{broken\_start}: a 4-tuple \texttt{(prompt, broken\_kernel, error\_log, native\_harness)}.
The fixer reads this 4-tuple as its full input on every iteration and emits a candidate patched solution.
Three invariants hold by construction: (1) \emph{reproducibility} (the harness produces the same failure from \texttt{broken\_kernel} every time), (2) \emph{cross-fixer fairness} (every fixer starts from the identical \texttt{(broken\_kernel, error\_log)} pair), and (3) \emph{debug-only} (no fresh kernel generation or significant performance improvement is requested).
The fixer's repair attempts proceed under the loop described in \S\ref{sec:benchmark:apparatus}; \cref{fig:pipeline-iter-vs-repeated} contrasts it with the generate-then-iterate protocol of prior benchmarks.

\subsection{Corpus}
\label{sec:benchmark:corpus}

\begin{wrapfigure}{r}{0.65\linewidth}
\vspace{-1.2em}
\centering
\includegraphics[width=\linewidth]{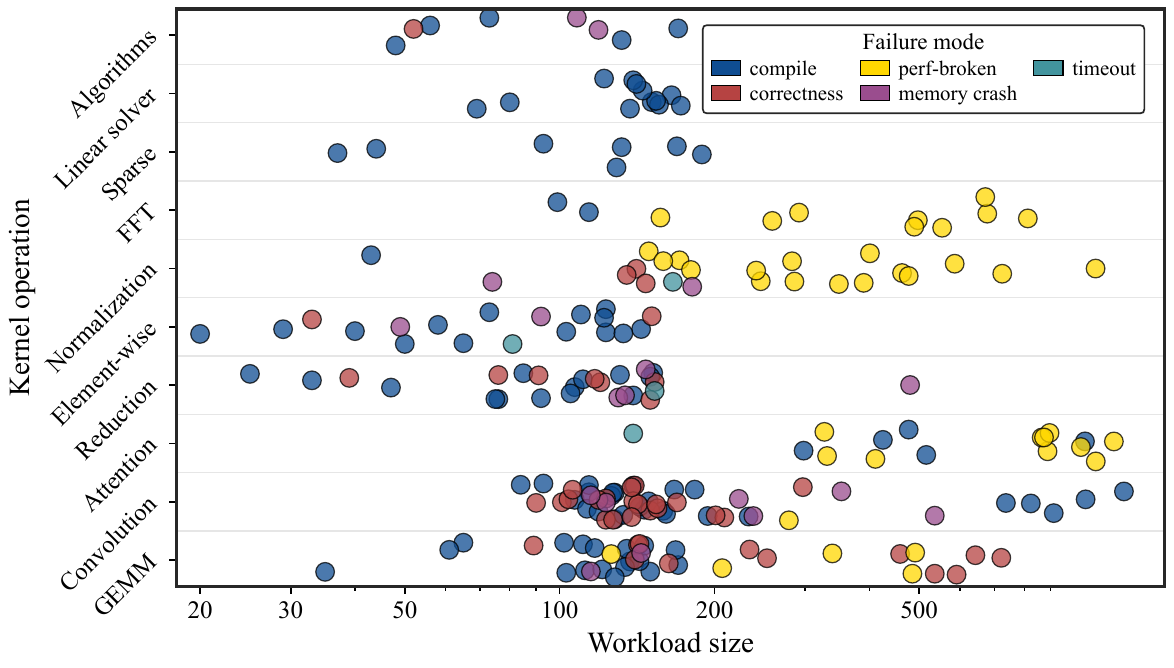}
\caption{Corpus coverage across workload size (lines of code, log scale, $x$), various domains ($y$), and 5 curation-time failure-modes (color, collapsed from the 8-category per-iteration taxonomy).}
\label{fig:corpus-map}
\end{wrapfigure}
The corpus contains \numtasks{} instances across existing generation benchmark KernelBench~\cite{ouyang2025kernelbench}, and our curated tasks where we instantiate the generation task first and then collect failure trajectories as final instances, spanning various high-quality CUDA implementations such as FlashAttention v2~\citep{fa2}, ThunderKittens~\citep{thunderkittensrepo}, and CUTLASS~\citep{cutlassrepo}. This design deliberately exercises debugging targeted at Hopper- and Blackwell-era kernels because some curated tasks rely on architecture-specific instructions. For example, the FP8 Tensor Core instructions required in task \texttt{Blackwell\_FP8\_GEMM} are only usable on Blackwell GPUs, making our tasks and evaluation hardware-aware. 

Candidate broken-starts are sourced from six external models and balanced across five error-category buckets, a curation-time collapse of the \numcategories{}-category per-iteration classifier (counts in \cref{tab:families-app,tab:sources-app,tab:errors-app}; classifier mapping in \cref{tab:taxonomy-app}). Each candidate broken-start is required to be (1) \emph{reproducible} (the failure reliably reproduces under the harness), (2) \emph{solvable} (at least one panel model produced a correct repair), (3) \emph{non-trivial} (not a one-character typo, judged by patch size), and (4) \emph{diverse} in error category. Each task stem is annotated with an empirical difficulty tier (L1--L5, full distribution in Appendix~\ref{app:difficulty}).  \Cref{fig:corpus-map} visualize this: in the joint (kernel-size, kernel-operation, failure-mode) layout, instances span 20--450 lines of code, cover all 10 operation types and all 5 failure modes, and leave no operation $\times$ failure-mode cell empty. The $L_5$ share ($\sim 44\%$, \cref{tab:tier-pass5}) is intentional: it preserves headroom for stronger future models.

\subsection{Evaluation setting}
\label{sec:benchmark:apparatus}

\paragraph{Seven fixers.}
The seven fixers (five cloud-API, two local vLLM-served; four also broken-start sources, using the asymmetric \passk{} of \cref{eq:asymmetric-passk}; full backend and axis coverage in Appendix~\ref{tab:committee}). Following the Bayesian critique of \citet{hariri2025don}, we treat Kendall $\tau$ as a panel-level statistic: each value summarizes the rank correlation between two six-element fixer-ranking vectors, while task-level uncertainty is already absorbed into each fixer's pass@5 estimate. Accordingly, we use $\tau$ as descriptive ordinal-change evidence rather than as a formal hypothesis test. Our grounded readings are the closed-form $z$-criterion from \S\ref{sec:benchmark:sensitivity} for declaring two-fixer flips resolved and the directly countable rank movements in \cref{tab:ranking}, where bold cells make every cross-axis flip explicit.

\paragraph{Debug loop.}
Each (fixer, task) loop runs up to $K = 5$ iterations. Iteration~$t$ takes the broken-start, error signal, and last $H = 4$ (candidate, feedback) pairs; the fixer emits a candidate; the harness compiles and runs it under the appropriate backend (Appendix~\ref{app:backends}); a deterministic classifier assigns one of \numcategories{} categories. The loop terminates on \texttt{passed} or on any of the four stagnation signals defined in \S\ref{sec:benchmark:metrics}; otherwise an error-aware feedback message at the configured level continues the loop.

\subsection{Pass@k as a Function of $(M, C, A)$: the Protocol-Conditional Envelope}
\label{sec:benchmark:sensitivity}

Prior CUDA benchmarks report \passk{}~\citep{chen2021evaluating} as a single number per (model, corpus); multi-prompt and leaderboard-sensitivity studies show this is brittle~\citep{mizrahi2024state,sclar2023quantifying,alzahrani2024benchmarks}, and a Bayesian re-analysis~\citep{hariri2025don} argues that finite sampling budgets cause non-trivial ranking instability.
We call the \emph{protocol-conditional envelope} of $\text{pass@}k(M, C, A)$, the joint characterization of per-axis swing $\Delta$ (protocol-induced variation) and cross-model Kendall $\tau$ (ranking stability), instantiating the variance- and difference-based perspectives of classical sensitivity analysis~\citep{saltelli2008global}.
We make this dependence explicit:
\begin{equation}
  \textit{pass@k}\bigl(M,\; C,\; A\bigr),
\label{eq:passk-as-function}
\end{equation}
where $M$ is the fixer, $C$ the corpus, and $A=(A_1,A_2,A_3,A_4)$ the \emph{evaluation-axis vector}, with performance-gate threshold, sampling strategy, feedback richness, and history depth, detailed below.
\paragraph{Coupling and measurement design.} The $(M,C,A)$ arguments are coupled in practice: $C$ is constrained by the source-model panel in the broken-start protocol; $A_3$ is gated on $M$'s backend because $L_4$ profiler signals require a local vLLM instance unavailable in cloud-API settings; and $A_1$'s performance-gate semantics depend on $C$'s reference-baseline definition. These couplings are features, not bugs, of apparatus-grounded evaluation and motivate the asymmetric pass@k weighting in \cref{eq:asymmetric-passk}. All results in \cref{sec:evaluation} therefore use a one-at-a-time (OAT) design~\citep{saltelli2008global}, sweeping each axis while holding the others at defaults; a full $9\times2\times5\times4=360$ cells per fixer is intractable at our 7-fixer scale. 

\paragraph{The four axes.}
We restrict attention to four axes central to prior CUDA-coding benchmark literature: (1) $A_1$, the \emph{performance-gate threshold} $p$ in $\textit{pass} = \textit{correctness} \;\wedge\; \textit{speedup} \geq p$, swept over $\{0.0, 0.1, 0.3, 0.5, 0.7, 0.9, 1.0, 1.5\}$ with default $p=0.7$ (prior work anchors at $p \in \{1.0, 2.0\}$~\citep{ouyang2025kernelbench,computeeval2025}); (2) $A_2$, the \emph{test-time sampling strategy} drawn from $\{\textit{iterative}, \textit{repeated}\}$ at the same five-call budget, default \emph{iterative} (iterative loops dominate debugging benchmarks versus single-shot evaluation; test-time compute is itself a load-bearing argument; (3) $A_3$, the \emph{feedback richness} $L \in \{L_0, L_1, L_2, L_3, L_4\}$ from no feedback to full per-category diagnostics, default $L_3$ (the richest level not requiring profilers unavailable in cloud-API contexts); (4) $A_4$, the \emph{conversation-history depth} $H \in \{1, 2, 3, 4\}$ rounds retained, default $H = 4$.\footnote{Because $H = 4$ is the upper bound of the sweep, $\textit{swing}(M, C, A_4)$ captures only the cost of reducing $H$ below the default.}

\paragraph{Quantification 1: swing.}
The single-axis swing addresses how much an adversary could shift a reported \passk{} by choosing a different default for one axis while holding the others fixed; the difference-based sensitivity measure of classical sensitivity analysis~\citep{saltelli2008global} applied to evaluation-protocol axes.
\begin{equation}
  \textit{swing}(M, C, A_i) \;=\; \max_{a_i} \textit{pass@k}(M, C, A) \;-\; \min_{a_i} \textit{pass@k}(M, C, A),
\label{eq:swing}
\end{equation}
with all axes other than $A_i$ held at their defaults.
Large swing on $A_i$ means that axis alone can move the headline number substantially, regardless of model quality. Following \citet{sclar2023quantifying}, who observe that performance spread over a sampled grid of protocol settings is a \emph{lower bound} on true sensitivity, our $\textit{swing}(M, C, A_i)$ likewise lower-bounds the true protocol-induced range:
\begin{equation}
  \textit{swing}(M, C, A_i) \;\le\; \sup_{a_i \in \mathcal{A}_i^{\text{full}}} \textit{pass@}k(M,C,A) \;-\; \inf_{a_i \in \mathcal{A}_i^{\text{full}}} \textit{pass@}k(M,C,A).
\label{eq:swing-lower-bound}
\end{equation}
Thus, the reported swing is a conservative lower bound on the full protocol-induced variation.

\paragraph{Quantification 2: Kendall $\tau$.}
Kendall $\tau$ between model-ranking vectors at two axis settings addresses whether a different reasonable protocol choice can put a different model at the top, regardless of how much any single number moved. Prior work shows that single-protocol leaderboards can reorder under modest evaluation-argument changes~\citep{alzahrani2024benchmarks,hariri2025don}, motivating an explicit ranking-stability metric alongside swing.
\begin{equation}
  \tau_{(A_i, a_i),(A_j, a_j)} \;=\; \textit{Kendall}\Bigl(\textit{rank}_M(A_i = a_i),\; \textit{rank}_M(A_j = a_j)\Bigr).
\label{eq:kendall}
\end{equation}
$\tau$ near $1$ means the cross-model ranking is robust to the axis change; $\tau$ near $0$ or negative means the ranking flips.
Swing and $\tau$ are not redundant: small swings can produce ranking flips when models cluster tightly, and large swings can preserve rank when all models shift together.
Both metrics are therefore necessary to characterise the protocol-conditional envelope fully.
Our use of $\tau$ as descriptive ordinal-change evidence finds explicit theoretical support in \citet{hariri2025don}: under a Dirichlet posterior, two models' observed ordering matches their true ordering with probability $\rho = \tfrac{1}{2}\bigl(1 + \mathrm{erf}(z/\sqrt{2})\bigr)$, $z = |\mu - \mu'|/\sqrt{\sigma^2 + (\sigma')^2}$, giving a closed-form criterion ($z \ge 1.645$) for declaring individual ranking flips resolved. Applying that criterion on a concrete panel requires per-cell variance estimates, so we cite it here as a principled tool that future panel extensions and replication studies should adopt rather than as a computation performed in the present study.


\subsection{Metric Family}
\label{sec:benchmark:metrics}

Beyond the headline \passk{}, we propose two complementary diagnostic additions: four stagnation signals that decompose ``budget burned without progress'' into interpretable modes, and a pass@1\,/\,debug\_rate decomposition that separates generation-strong from debug-strong fixer profiles. In particular, \emph{repair by degeneration} is surfaced by the \texttt{duplicate\_code} and \texttt{no\_progress} signals, which capture degenerate iteration patterns rather than productive repair.

\paragraph{Stagnation signals.}
\passk{} captures whether the task passes within $k$ iterations but not whether each iteration does real work. The four signals below provide a deterministic decomposition of ``the model wastes its budget''; each both terminates the loop early and is reported per-task as an independent quality measurement (\cref{sec:eval:stagnation}). These signals draw on three prior-art lines: cycle detection in iterative search \citep{knuth2014art,brent1980improved,flajolet1989random}, stall generations in evolutionary algorithms \citep{de1975analysis}, and degeneration in LLM text generation \citep{holtzman2019curious}. Since iterative LLM refinement is not guaranteed to converge~\citep{madaan2023selfrefineiterativerefinementselffeedback,olausson2023self}, we apply four explicit non-convergence signals calibrated to the $K = 5$ budget so that each can fire at least once on a stagnating trajectory without triggering on any healthy one. A threshold-sensitivity ablation is in Appendix~\ref{app:stagnation-thresholds}.

The four signals are (1) \texttt{duplicate\_code} (model resubmits its own previous attempt verbatim; a 1-step instance of \emph{degeneration}), (2) \texttt{code\_cycle} (model re-tries a strictly earlier rejected solution; a Floyd--Brent cycle in the solution trajectory), (3) \texttt{category\_oscillation} (model alternates between failure modes rather than narrowing the fix space), and (4) \texttt{no\_progress} (model stuck on the same concrete bug; \emph{stall iterations}).
Closed-form bounds for \texttt{code\_cycle}, \texttt{duplicate\_code}, and \texttt{no\_progress} are derived in Appendix~\ref{app:stagnation-grounding}; \texttt{category\_oscillation} is an empirical companion signal without closed-form bound. The four signals are jointly orthogonal to \passk{}: a high-\passk{} model can still waste its budget under \texttt{no\_progress}, and a high-stagnation model can still solve a non-trivial fraction of the corpus (\cref{sec:eval:stagnation}).

\paragraph{Decomposition into pass@1 and debug rate.}
Two fixers can share an identical \passk{} headline while differing sharply in \emph{how} they reach it. To expose this, we report pass@1 and \texttt{debug\_rate@k} separately alongside the combined figure. The two quantities are linked by
\begin{equation*}
  \textit{pass@k} \;=\; \textit{pass@1} \;+\; (1 - \textit{pass@1}) \cdot \textit{debug\_rate@k},
\end{equation*}
where \texttt{debug\_rate@k} is the fraction of tasks that failed on iteration~1 but passed within $k$ iterations. A generation-strong\,/\,debug-weak profile exhibits high pass@1 and low \texttt{debug\_rate}; a generation-weak\,/\,debug-strong profile exhibits the reverse. Prior CUDA benchmarks report only the combined headline and cannot make this distinction at any single axis setting.

%% file: sections/experiments.tex
\section{Evaluation}
\label{sec:evaluation}

This section evaluates how seven frontier LLMs debug the \numtasks{} broken-start CUDA tasks in \benchmark{}. We treat $\text{pass@}k(M,C,A)$ as the object being measured, where $M$ is the fixer, $C$ is \benchmark{}, and $A$ specifies the protocol axes introduced in \S\ref{sec:benchmark:sensitivity}. The experiments are conducted on NVIDIA RTX PRO 6000 GPU (Blackwell) and H200 GPU (Hopper), according to the specific hardware requirement of each task.

\textbf{Models and default protocol.} Evaluated models consists of GPT-5.4~\citep{singh2026openaigpt5card}, Qwen3.6-Plus and Qwen3.6-27B~\citep{yang2025qwen3technicalreport}, Gemma-4-31B-it~\citep{gemmateam2025gemma3technicalreport}, Kimi-k2.6~\citep{kimiteam2026kimik2openagentic}, GLM-4.7~\citep{5team2025glm45agenticreasoningcoding}, and MiniMax-M2.7~\citep{minimax2025minimaxm1scalingtesttimecompute}. Unless otherwise stated, each fixer receives up to $K{=}5$ repair attempts with $H{=}4$ retained rounds, feedback level $L_3$, iterative sampling at $T{=}0.7$, and performance gate $p{=}0.7$. Scores use the asymmetric \passk{} rule in \cref{eq:asymmetric-passk}; full backend, source-overlap, and axis-coverage details are provided in Appendix~\ref{app:fixers}.

\textbf{Protocol sweeps.} Starting from the default protocol, we vary one axis at a time and hold the others fixed: performance gate $A_1$, sampling strategy $A_2$, feedback richness $A_3$, and conversation-history depth $A_4$. For each sweep we report score swing to quantify absolute sensitivity, and Kendall's $\tau$ where applicable to quantify cross-model ranking stability. This design isolates which evaluation choices make models look stronger, weaker, or differently ranked, while avoiding an intractable full Cartesian sweep over all protocol combinations.

\subsection{Performance-Gate Threshold $A_1$}
\label{sec:eval:p}

Raising the performance-gate $p$ from $0$ to $1.5$ reduces \passk{} by 8.5--40.0 percentage points across the seven models. The largest drop is on GPT-5.4, from $55.5\%$ to $15.5\%$, and the smallest is on MiniMax-M2.7, from $13.5\%$ to $5.0\%$; the full threshold sweep is provided in Appendix~\ref{app:per-axis-raw}.

\begin{figure}[t]
\centering
\begin{minipage}[c]{0.50\linewidth}
\centering
\small
\setlength{\tabcolsep}{4pt}
\renewcommand{\arraystretch}{1.05}
\begin{tabular}{lccc}
\toprule
Model & iter-1 med. & iter-N med. & deg. rate \\
\midrule
GPT-5.4       & $1.53\times$ & $1.10\times$ & $36.8\%$ \\
Qwen3.6-Plus  & $0.94\times$ & $1.13\times$ & $45.5\%$ \\
Gemma-4-31B-it   & $1.03\times$ & $0.95\times$ & $56.2\%$ \\
Qwen3.6-27B   & $1.03\times$ & $1.49\times$ & $25.0\%$ \\
Kimi-k2.6     & $1.07\times$ & $1.25\times$ & $15.4\%$ \\
GLM-4.7       & $1.04\times$ & $1.93\times$ & $\phantom{0}9.1\%$ \\
MiniMax-M2.7  & $1.03\times$ & $1.63\times$ & $35.7\%$ \\
\bottomrule
\end{tabular}
\end{minipage}\hfill
\begin{minipage}[c]{0.48\linewidth}
\centering
\includegraphics[width=\linewidth]{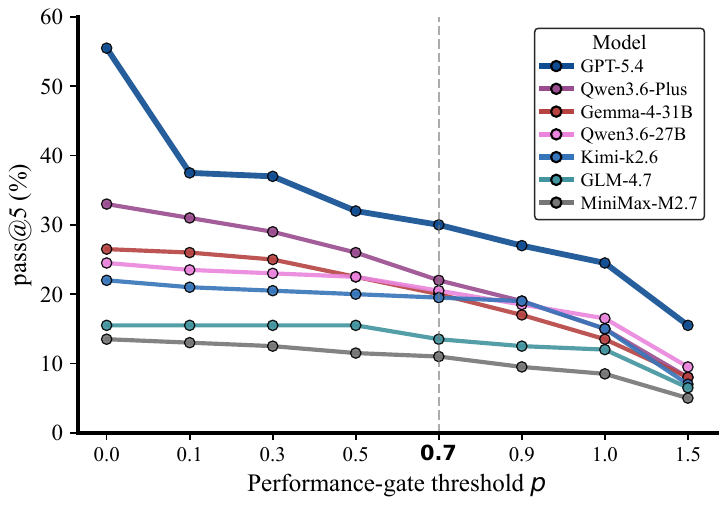}
\end{minipage}
\caption{(Left) Per-model degeneration evidence on the corpus's eventually-passing tasks: median speedup of iter-1 passes vs.\ iter-N ($N{>}1$) passes, and the degeneration rate (share of iter-N passes whose speedup falls below the reference baseline, $\textit{i.e.}\ {<}1.0\times$). (Right) Per-model pass@5 vs.\ performance-gate $p$. Dashed line marks the default $p{=}0.7$.}
\label{fig:f3-a1}
\end{figure}
This gate choice also changes how much separation the benchmark exposes between models: the absolute gap between strongest and weakest models contracts from ${\sim}42.0$ percentage points at $p=0.0$ to ${\sim}10.5$ at $p=1.5$. Benchmarks fixing a single operating point thus anchor leaderboards to one slice of a wider curve. Speedup ratios use 10-launch post hoc re-measurement, with CV reported at a 3\% threshold; Appendix~\ref{app:timing} gives the protocol.

Interpreting a lower \passk{} as degeneration would be premature, since the same pattern could arise from weaker iter-1 candidates, stronger reference baselines, or real degenerative repair. The iter-1 versus iter-N speedup comparison in the left panel of \cref{fig:f3-a1} isolates one component of this ambiguity. We restrict to tasks each model solves at $p=0$ and compare the model's own iter-N pass to its own iter-1 pass. Such comparison controls for both reference-baseline strength and cross-model generation gap, leaving within-model temporal degradation as the residual signal. Per-model degeneration rates range from 9.1\% to 56.2\%, and roughly one third of iter-N passes across evaluated models fall below the reference baseline (i.e., $<1.0\times$ speedup); $p=0$ still credits them, whereas $p \geq 1.0$ does not.

Cross-model rank order is largely preserved within each $p$ bucket. GPT-5.4 remains the strongest model across thresholds, even as margins compress sharply. Swing magnitude and ranking stability therefore decouple: a benchmark can be highly sensitive to $p$ in absolute \passk{} while remaining comparatively stable in relative ordering. We formalize this in \S\ref{sec:eval:cross}.

\subsection{Test-Time Sampling Strategy $A_2$}
\label{sec:eval:method}

At the same five-call budget, iterative repair and repeated sampling yield qualitatively different cross-model rankings. The directional pattern is mixed: 4 of 7 models favor iteration, with gains up to +9.9 percentage points on Qwen3.6-Plus; 2 favor repeated sampling, with the largest iteration deficit at -5.6 percentage points on GPT-5.4; and Gemma-4-31B-it is effectively unchanged. Thus, how the same call budget is spent can change which models benefit, not merely the average score. Full per-model values are provided in Appendix~\ref{app:per-axis-raw}; \cref{fig:pipeline-iter-vs-repeated} illustrates the two sampling protocols.

The dominant SWE-bench-style agent loop is iterative by construction \citep{jimenez2023swe, baronio2025kevin}. Under a fixed \passk{} budget, that design choice not only shifts absolute performance: it commits to one particular cross-model ordering. Evaluating the same systems with repeated sampling at the same budget substantially reconfigures that ordering, reversing several pairwise relationships.

The Kendall $\tau = 0.47$ result in \cref{tab:ranking} identifies test-time sampling strategy as the strongest cross-axis leaderboard reshuffle in the study. Notably, the largest model-level swing on this axis is only 9.9 percentage points (Qwen3.6-Plus), yet the induced ordering changes exceed those of every other axis. Swing magnitude and ranking stability are thus distinct quantities, as we discussed in \S\ref{sec:eval:cross}.

\subsection{Feedback Richness $A_3$ and Conversation-History Depth $A_4$}
\label{sec:eval:feedback}

\begin{figure}[t]
\vspace{-0.5em}
\centering
\includegraphics[width=\linewidth]{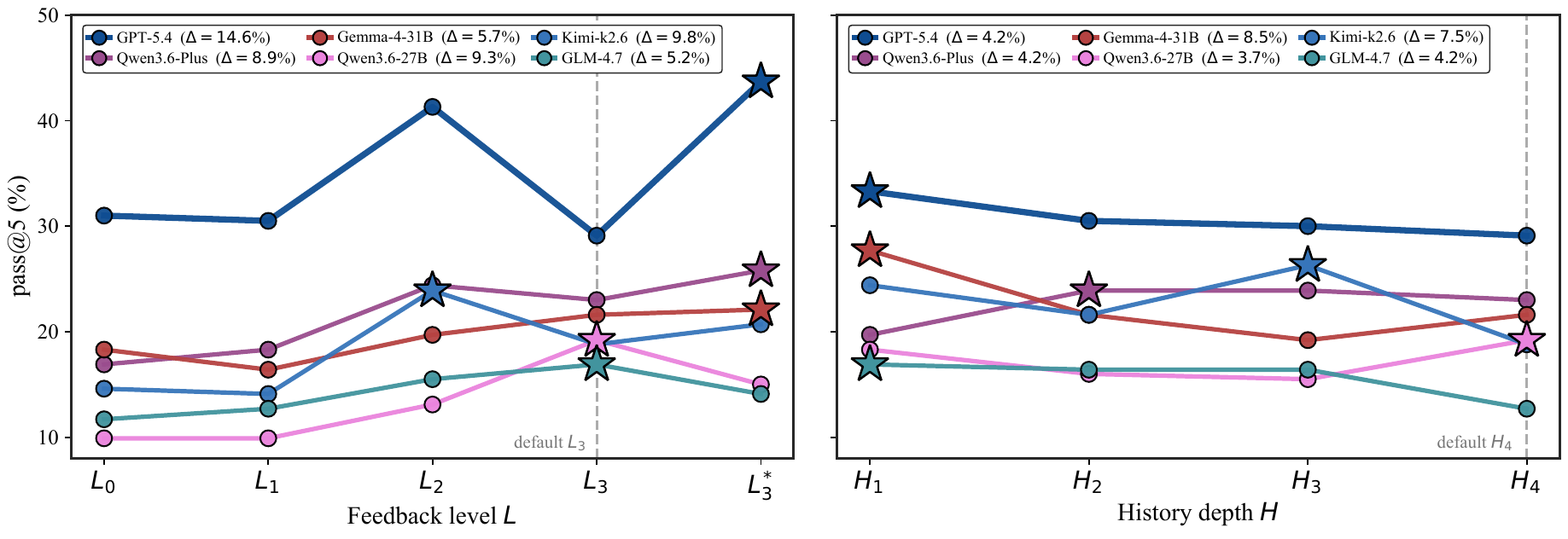}
\vspace{-0.5em}
\caption{\textbf{Feedback richness $A_3$ (left) and conversation-history depth $A_4$ (right).} Per-model \passk{} trajectories; star marks each model's best setting, dashed grey line marks the default.}
\label{fig:axes-a3a4}
\vspace{-1.2em}
\end{figure}

\textbf{Feedback richness $A_3$.} The \passk{} sweep over feedback-richness levels shows substantial but uneven sensitivity, with per-model pass@5 ranges spanning 3.8--12.7 percentage points. GPT-5.4 exhibits the largest response, improving from 31.0\% at $L0$ to 43.7\% at $L3^\ast$ for a gain of 12.7 points. At the same time, the best feedback level is not uniform: Kimi is a clear non-monotone counter-example, peaking at $L2$ and degrading at richer levels. This pattern argues against evaluating only a single fixed feedback setting, as is common in prior works \citep{peng2025perfcodegen, xia2023keep}. Full per-level values are provided in Appendix~\ref{app:per-axis-raw}.

\textbf{Conversation-history depth $A_4$.} The history-depth sweep is smaller on average but still materially consequential, with per-model pass@5 ranges of 0.9--8.5 percentage points. No single history budget dominates: the best $H$ is model-specific. Gemma-4-31B-it shows the largest swing, reaching 27.7\% at $H1$ but dropping to 19.2\% at $H3$, an 8.5-point spread. As with $A_3$, these results caution against max-budget defaults often adopted in prior repair studies \citep{xia2023keep, baronio2025kevin}. Full per-history values are provided in Appendix~\ref{app:per-axis-raw}.

\textbf{Ranking stability under both axes.} Despite these within-model swings, cross-model ordering is more stable than raw deltas suggest for $A_3$: $\tau_{A_3}=0.73$, indicating near-preservation of the cross-model ranking, with only Kimi notably rising from rank 5 to rank 3. For $A_4$, $\tau_{A_4}=0.60$ reflects a more moderate reshuffle. Both $A_3$ and $A_4$ thus provide independent decouplings of swing magnitude from ranking stability, a separation we return to in \S\ref{sec:eval:cross}.

\subsection{Cross-Axis Ranking Robustness}
\label{sec:eval:cross}

\textbf{Phenomenology.} Across the four \benchmark{} axis settings, the extremes of the leaderboard are invariant: GPT-5.4 remains rank~1 and GLM-4.7 remains rank~6 under every protocol. Instability is concentrated in the interior, where middle positions reshuffle materially once the default iterative setting is replaced by compute-matched alternatives. We direct the reader to \cref{tab:ranking}, where bold cells make cross-axis flips visually explicit.

\begin{wraptable}{r}{0.55\linewidth}
\vspace{-1.2em}
    \centering
    \scriptsize
    \setlength{\tabcolsep}{3pt}
    \renewcommand{\arraystretch}{1.05}
    \begin{tabular}{lllll}
        \toprule
        Rank & Default iter & Repeated  & Best-fb  & Best-hist \\
        \midrule
        1 & GPT-5.4         & GPT-5.4              & GPT-5.4              & GPT-5.4         \\
        2 & Qwen3.6-Plus    & \textbf{Gemma-4}     & Qwen3.6-Plus         & \textbf{Gemma-4}\\
        3 & Gemma-4         & \textbf{Kimi}        & \textbf{Kimi}        & \textbf{Kimi}   \\
        4 & Qwen3.6-27B     & Qwen3.6-27B          & \textbf{Gemma-4}     & Qwen3.6-Plus    \\
        5 & Kimi            & \textbf{Qwen3.6-Plus}& Qwen3.6-27B          & Qwen3.6-27B     \\
        6 & GLM-4.7         & GLM-4.7              & GLM-4.7              & GLM-4.7         \\
        \midrule
        \multicolumn{5}{c}{Kendall $\tau$ vs.\ default} \\
        \midrule
        & 1.00 & 0.47 & 0.73 & 0.60 \\
        \bottomrule
    \end{tabular}
    \caption{Cross-model ranking under axes $A_2$, $A_3$, $A_4$; bold = flip vs.\ default. Six fixers.}
    \label{tab:ranking}
\vspace{-1.5em}
\end{wraptable}
\textbf{Translating $\tau$ to swaps.} These swap counts are grounded in directly enumerable rank movements from \cref{tab:ranking} and therefore do not depend on interpreting $\tau$ as a population estimator. Relative to the iterative default, all three non-default axes induce ordinal reshuffles of clearly different severity. On a six-model ranking, $\tau = 0.47$ corresponds to roughly four adjacent-rank swaps, $\tau = 0.60$ to about three, and $\tau = 0.73$ to approximately two. The method variation thus produces the strongest reordering, history variation a moderate one, and feedback variation a near-preserving one.

\textbf{Decoupling synthesised.} The four axes show that absolute performance swing and ranking stability do not move in lockstep under \benchmark{}. Swing ranges from 9.9 percentage points on $A_2$ to 40.0 on $A_1$, while the $\tau$ span runs from the strongest reshuffle on $A_2$ to near preservation on $A_3$, so the smallest-swing axis induces the largest ordinal disruption, and a larger-swing axis can leave the leaderboard largely unchanged. Three of the four axes thus reshape the \passk{} ordering at matched effective compute, meaning any single default protocol commits the study to one specific cross-model ranking among multiple defensible alternatives.

\subsection{Metric Family Snapshot}
\label{sec:eval:stagnation}


\textbf{Generation-strong vs.\ debug-strong split.} \texttt{debug\_rate@5} ranges from $5.6\%$ (GLM-4.7) to $17.1\%$ (GPT-5.4). GPT-5.4 is the only fixer whose \texttt{debug\_rate@5} exceeds its pass@1, making it the only fixer that recovers measurably from iter-1 failures rather than relying on iter-1 success.

\textbf{Dominant stagnation signals split the fixers.} Four fixers (GPT-5.4, Gemma-4-31B-it, Qwen3.6-Plus, Kimi-k2.6) are dominated by \texttt{no\_progress}, indicating repeated attempts stuck on the same concrete bug across iterations. Two others (Qwen3.6-27B, GLM-4.7) are dominated by \texttt{duplicate\_code}, indicating verbatim resubmission of an earlier attempt. These are qualitatively different failure modes that headline \passk{} collapses into a single aggregate number.

\textbf{Default-protocol commitment.} Six of seven fixers prefer a non-default protocol on at least one axis, and per-bucket \texttt{fix\_rate} varies sharply within each fixer (13--30 pp spread; GPT-5.4: 47.5\% \texttt{compile} vs.\ 70.7\% \texttt{perf-broken}). Full data on this are provided in Appendix~\ref{app:per-axis-raw}.

%% file: sections/conclusion.tex
\section{Conclusion}
\label{sec:conclusion}

This paper identifies CUDA debugging as an overlooked but critical capability for LLM-based automated CUDA programming. We introduce \benchmark{} to evaluate this capability. Proposed $\text{pass@}k(M,C,A)$ makes the fixer, corpus, and evaluation axes explicit rather than hiding them inside a single aggregate \passk{}. This protocol-aware evaluation exposes \emph{repair by degeneration}, where fixers pass correctness tests by replacing optimized code with slower fallbacks. The evaluation of seven LLMs on \numtasks{} tasks shows that stricter performance requirements reduce reported success by up to 40.0 percentage points, and reasonable protocol choices can also change model rankings.

%% file: sections/appendix.tex
\appendix

\section{Asymmetric \passk{} Normalisation}
\label{sec:benchmark:asymm}

When the fixer is also a broken-start source, its pipeline is structurally one step ahead: the broken-start is, in effect, the fixer's own iteration~1, so our pipeline's iteration~1 already sits at what would be the fixer's iteration~2 of an implicit extended sequence.
We adopt an asymmetric \passk{} as a conservative attempt-budget equalisation that prevents this implicit head-start from accumulating into the headline figure, independently of whether self-source overlap empirically biases scores in any given panel.
For cross-source instances we report \texttt{pass@5} over the fixer's iterations 1--5; for self-source instances we report \texttt{pass@4} by dropping the fifth iteration.
Formally, with $\textit{src}(t)$ denoting the source model of task~$t$,
\begin{equation}
\textit{pass}^*(M, t) \;=\;
\begin{cases}
  \text{pass@5}(M, t), & \text{if } M \neq \textit{src}(t),\\
  \text{pass@4}(M, t), & \text{if } M = \textit{src}(t).
\end{cases}
\label{eq:asymmetric-passk}
\end{equation}
This rule equalises the \emph{effective} attempt budget across the two conditions without discarding any task.
Among our seven fixers, four (Kimi, Qwen3.6-Plus, Gemma-4-31B-it, GPT-5.4) are also source models and therefore use the asymmetric rule; the other three (Qwen3.6-27B, GLM-4.7, MiniMax-M2.7) always use \texttt{pass@5}.
We treat \cref{eq:asymmetric-passk} as a methodological default rather than a correction calibrated to a measured bias: any future panel with substantial source-fixer overlap should adopt the same rule even before measuring whether self-source overlap inflates scores in that panel.

\section{Fixers}
\label{app:fixers}

\Cref{tab:committee} tabulates the seven fixers referenced from \S\ref{sec:benchmark:apparatus}: per-model backend assignment, broken-start source overlap, and axis coverage.

\begin{table}[h]
    \centering
    \small
    \setlength{\tabcolsep}{8pt}
    \renewcommand{\arraystretch}{1.05}
    \begin{tabular}{lccc}
        \toprule
        Model & Backend & Source? & Axes covered \\
        \midrule
        Kimi-k2.6      & Cloud API   & \cmark & $A_1$--$A_4$ \\
        Qwen3.6-Plus   & Cloud API   & \cmark & $A_1$--$A_4$ \\
        Qwen3.6-27B    & Local vLLM  & \xmark & $A_1$--$A_4$ \\
        Gemma-4-31B-it & Local vLLM  & \cmark & $A_1$--$A_4$ \\
        GLM-4.7        & Cloud API   & \xmark & $A_1$--$A_4$ \\
        GPT-5.4        & Cloud API   & \cmark & $A_1$--$A_4$ \\
        MiniMax-M2.7   & Cloud API   & \xmark & $A_1$, $A_2$ \\
        \bottomrule
    \end{tabular}
    \caption{Seven fixers. \cmark{} = also a broken-start source (uses \cref{eq:asymmetric-passk}). All rows run under the default protocol ($K{=}5$, $H{=}4$, feedback $L_3$, performance-gate $p{=}0.7$, iterative $T{=}0.7$; $T{=}1.0$ for repeated sampling on $A_2$). MiniMax-M2.7 was added late in the evaluation cycle and only covers axes $A_1$ and $A_2$; the cross-axis Kendall $\tau$ analysis in \S\ref{sec:eval:cross} is therefore computed on the six-model subset with full axis coverage. As a sensitivity check, including MiniMax on $A_2$ (its only non-default axis) shifts $\tau_{A_2}$ from 0.47 (six-model) to 0.62 (seven-model), so the six-model figure overstates rather than understates instability. This is a panel-level analog of a sensitivity analysis: changing panel composition by one fixer perturbs $\tau$ within a controlled range, consistent with treating it as a descriptor of the specific panel rather than an estimator of an underlying population correlation.}
    \label{tab:committee}
\end{table}

\section{Implementation Details}
\label{app:backends}

\subsection{Task Schema and Execution Backends}

Each task directory exposes three artifacts. The \texttt{input/<stem>.json} file is the metadata contract consumed by every backend:

\begin{verbatim}
{
  "task_id":       "CUDA/10",
  "source":        "cuda_samples",
  "backend":       "nvcc",
  "solution_file": "solution.cu",
  "build_cmd":     "nvcc -O2 -arch=sm_80 -o test
                    test_main.cu solution.cu",
  "test_cmd":      "./test",
  "min_sm":        80,
  "requires":      [],
  "anti_cheat":    ["cublasSgemm", "cusolverDnSgetrf"],
  "timing_parser": "^Kernel time: ([0-9.]+) ms"
}
\end{verbatim}

\paragraph{Prompts.} \texttt{prompts/<stem>.txt} contains the task description. Prompts are written to specify the task and the test contract, not to hand the model the reference solution. Anti-cheat substrings are checked on the submitted solution at pre-flight.

\paragraph{Testbenches.} \texttt{testbench/<stem>/} holds the test harness, header stubs, and reference files. For KernelBench tasks this is the PyTorch reference module; for CUTLASS-pool tasks it is a Make-ready project subtree; for raw-\texttt{nvcc} tasks it is the test \texttt{.cu} and the required headers.

\subsection{Feedback Templates}

The feedback templates routed by the per-category dispatcher of \S\ref{sec:benchmark:apparatus} are short. We list them here for reproducibility; each template's body is the relevant fragment of the harness log, not model-generated summarisation.

\paragraph{Buildability.} \texttt{``Your previous solution did not compile. The first $N$ lines of \texttt{nvcc} stderr are:\textbackslash n<stderr>\textbackslash nPlease revise the solution to compile.''}

\paragraph{Integration.} \texttt{``Your previous solution compiles but does not match the expected harness contract: <contract diff>. Please revise the interface.''}

\paragraph{Illegal memory access.} \texttt{``Your previous solution triggered a runtime CUDA error: <error signature>. The last successful checkpoint was <test-log tail>. Please diagnose and revise.''}

\paragraph{Out of memory.} \texttt{``Your previous solution failed with \texttt{cudaMalloc} OOM on input <shape>. Please reduce memory footprint or tile more aggressively.''}

\paragraph{Timeout.} \texttt{``Your previous solution exceeded the $T$-second test budget on input <shape>. Please investigate for unbounded loops or low-occupancy launches.''}

\paragraph{Functional correctness.} \texttt{``Your previous solution compiled and ran but produced incorrect output on input <shape>. tolerance=<tol>, max abs error=<max>, mean abs error=<mean>, sample mismatches=<list>. Please revise.''}

\subsection{Two-Phase Protocol}

Phase 1 runs one iteration on every task in $\mathcal{T}$; let $\mathcal{F}$ denote the set of tasks that fail Phase 1. Phase 2 runs up to $K-1$ additional iterations on every task in $\mathcal{F}$, with each task's Phase-2 trajectory starting from the feedback of the Phase-1 failure. The concatenation of the Phase-1 iteration with the Phase-2 trajectory yields exactly the $K$-iteration trajectory the naive protocol would have produced for each task, so \passone, \passk, \debugk, per-category \texttt{fix\_rate}, and every quality signal coincide with the naive protocol by construction. Total call count is $N + (K-1)|\mathcal{F}|$, compared with $NK$ under the naive schedule; the relative saving grows with \passone{}.

\section{Formal Definitions}

\subsection{Stagnation Signals}

Four signals, evaluated after each iteration:

\begin{itemize}[leftmargin=1.5em]
    \item \textbf{\texttt{duplicate\_code}}: SHA-256 of the current solution matches the previous iteration's. Fires as soon as the duplicate is observed.
    \item \textbf{\texttt{code\_cycle}}: SHA-256 of the current solution matches any earlier iteration in the same task trajectory.
    \item \textbf{\texttt{category\_oscillation}}: At least three category transitions in the window of the last five iterations. Formally, $|\{t: \text{cat}(t) \neq \text{cat}(t-1), t\in\{\max(2,n-4),\dots,n\}\}|\geq 3$.
    \item \textbf{\texttt{no\_progress}}: The tuple (category, primary error signature) is unchanged across three consecutive iterations, where the primary error signature is the first matched error-pattern token from the harness log.
\end{itemize}

When any of these fires, the harness records the signal name as the task's \texttt{stop\_reason} and terminates the loop. The iteration on which the signal fires is included in the trajectory and counted against the iteration budget.

\subsection{Theoretical Grounding of Stagnation Signals}
\label{app:stagnation-grounding}

(i)~For \texttt{code\_cycle}, \citet{flajolet1989random} give the expected first-collision position $\sqrt{\pi n / 2}$ for random functional iteration on an $n$-state space; with SHA-256, $n = 2^{256}$, so under $K = 5$ the spurious-collision probability is bounded by $\binom{5}{2} \cdot 2^{-256} \approx 8.6 \times 10^{-77}$. Any detected cycle is therefore genuine model behaviour rather than hash collision, and \texttt{code\_cycle} has zero false-positive rate by construction.
(ii)~For \texttt{duplicate\_code}, \citet[Fig.~4]{holtzman2019curious} show that under maximum-likelihood decoding the probability of repeating a phrase increases monotonically with each repetition, creating a positive-feedback loop regardless of phrase length or content; our 1-step threshold is the minimal trigger for this established mechanism.
(iii)~For \texttt{no\_progress}, an invariant $(\text{category}, \text{error signature})$ tuple across $T$ iterations is a fixed point of the state-conditional feedback dynamics; $T = 3$ is the minimal detection window in $K = 5$ consistent with this fixed-point interpretation, paralleling stall-generation criteria in evolutionary algorithms \citep{de1975analysis}.
\texttt{category\_oscillation} is reported without closed-form bound and is best understood as an empirical companion signal.

\subsection{Metric Definitions}

Let $\mathcal{T}$ denote the evaluation set of size $N$, and let $t_i\in\{\text{pass},\text{fail}\}^{K_i}$ be the trajectory of task $i\in\mathcal{T}$ of length $K_i\leq K$. Let $\text{cat}(i,j)$ denote the category of iteration $j$ on task $i$.

\begin{align*}
\text{pass@}1            &= \frac{1}{N}\sum_{i\in\mathcal{T}} \mathbb{1}[t_{i,1}=\text{pass}] \\
\text{pass@}k            &= \frac{1}{N}\sum_{i\in\mathcal{T}} \mathbb{1}[\exists j\leq K,\, t_{i,j}=\text{pass}] \\
\text{debug\_rate@}k     &= \frac{|\{i : t_{i,1}=\text{fail}\wedge \exists j\leq K,\, t_{i,j}=\text{pass}\}|}{|\{i:t_{i,1}=\text{fail}\}|} \\
\text{fix\_rate}[c]      &= \frac{|\{i:\text{cat}(i,1)=c \wedge \exists j,\, t_{i,j}=\text{pass}\}|}{|\{i:\text{cat}(i,1)=c\}|}
\end{align*}

\texttt{stagnation\_rate} is the fraction of tasks whose \texttt{stop\_reason} is any of the four stagnation signals. \texttt{oscillation\_rate} is the fraction whose stop reason is \texttt{category\_oscillation} specifically. \texttt{progression\_rate} is the fraction of failing tasks whose trajectory contains at least one category transition in a monotone severity order (\texttt{buildability}/\texttt{integration}/\texttt{environment\_dependency} $\rightarrow$ \texttt{out\_of\_memory}/\texttt{illegal\_memory\_access}/\texttt{timeout} $\rightarrow$ \texttt{functional\_correctness} $\rightarrow$ \texttt{passed}). \texttt{unique\_approach\_ratio} is the count of distinct solution code hashes divided by the total iteration count.

The error-transition matrix $M\in\mathbb{R}^{8\times 8}$ has entries $M_{a,b}=\Pr(\text{cat}(i,j+1)=b \mid \text{cat}(i,j)=a)$ estimated from the iteration-pair empirical distribution.

\subsection{Difficulty Stratification}
\label{app:difficulty}

Each task stem is annotated with an empirical difficulty tier $L_1$--$L_5$ derived from the iteration count required for the six fixers (excluding MiniMax-M2.7) to reach a passing solution. $L_1$: solved by all six members with mean iter\;$\leq 2$; $L_2$: same but mean iter\;$> 2$; $L_3$: solved by 3--5 of the six; $L_4$: solved by 1--2; $L_5$: not solved by any. \Cref{tab:tier-pass5} reports the per-fixer \passk{} stratified by tier under the default protocol; the empirical tier distribution on the corpus is $L_1{=}12$, $L_2{=}0$, $L_3{=}48$, $L_4{=}57$, $L_5{=}91$ ($n{=}208$; the remaining 5 tasks fall outside the six fixers' manifest coverage). The empty $L_2$ row is itself a finding: tasks solvable by all six members are uniformly solved on the first iteration by all of them, suggesting a binary easy/hard structure rather than a smooth difficulty gradient.

\begin{table}[H]
    \centering
    \small
    \setlength{\tabcolsep}{6pt}
    \renewcommand{\arraystretch}{1.05}
    \begin{tabular}{lccccc}
        \toprule
        Model & $L_1$ ($n{=}12$) & $L_2$ ($n{=}0$) & $L_3$ ($n{=}48$) & $L_4$ ($n{=}57$) & $L_5$ ($n{=}91$) \\
        \midrule
        GPT-5.4       & 100.0\% & --- & 93.8\% & 94.7\% & 0.0\% \\
        Qwen3.6-Plus  & 100.0\% & --- & 89.6\% & 19.3\% & 0.0\% \\
        Gemma-4-31B-it   & 100.0\% & --- & 68.8\% & 14.0\% & 0.0\% \\
        Qwen3.6-27B   & 100.0\% & --- & 72.9\% &  3.5\% & 0.0\% \\
        Kimi-k2.6     & 100.0\% & --- & 56.2\% &  8.8\% & 0.0\% \\
        GLM-4.7       & 100.0\% & --- & 39.6\% &  0.0\% & 0.0\% \\
        MiniMax-M2.7  &  50.0\% & --- & 37.5\% &  5.3\% & 0.0\% \\
        \bottomrule
    \end{tabular}
    \caption{Per-fixer \passk{} stratified by difficulty tier at the default protocol ($p{=}0.7$). Tiers are computed over the six fixers (excluding MiniMax-M2.7); MiniMax row is reported on the same tier partition for comparability. $L_5$ tasks are unsolved by every fixer in the tier-induction set, so all fixers register $0\%$ on this slice by construction; $L_5$ thus measures residual solvability beyond the tier-induction set, which is empirically zero on this corpus. MiniMax's $L_1$ row (50\%) is therefore not anomalous: $L_1$ membership is defined relative to the six-model tier-induction set, not MiniMax.}
    \label{tab:tier-pass5}
\end{table}

\section{Error Taxonomy (Eight-Category Schema)}
\label{app:taxonomy}

The per-iteration classifier assigns one of \numcategories{} fine-grained labels listed in \cref{tab:taxonomy-app}. The five-bucket collapse used for broken-start curation in \cref{tab:errors-app} groups categories~1--3 into \texttt{compile\_error}, categories~4--5 into \texttt{memory\_crash}, with \texttt{timeout} (6) and \texttt{functional\_correctness} (7) renamed to \texttt{logic\_error}, and \texttt{passed} (8) standing alone.

\begin{table}[h]
    \centering
    \small
    \setlength{\tabcolsep}{5pt}
    \renewcommand{\arraystretch}{1.05}
    \begin{tabularx}{\linewidth}{>{\centering\arraybackslash}p{0.4cm} >{\raggedright\arraybackslash}p{4.0cm} >{\raggedright\arraybackslash}p{2.4cm} >{\raggedright\arraybackslash}X}
        \toprule
        \# & Category (8-cat) & 5-bucket & Signal \\
        \midrule
        1 & \texttt{environment\_dependency} & \texttt{compile\_error} & GPU architecture mismatch, missing libraries \\
        2 & \texttt{integration} & \texttt{compile\_error} & Harness-contract mismatch \\
        3 & \texttt{buildability} & \texttt{compile\_error} & Syntax, type, or API errors \\
        4 & \texttt{out\_of\_memory} & \texttt{memory\_crash} & \texttt{cudaMalloc} OOM \\
        5 & \texttt{illegal\_memory\_access} & \texttt{memory\_crash} & Segfault, \texttt{cudaErrorIllegalAddress} \\
        6 & \texttt{timeout} & \texttt{timeout} & Execution exceeds wall-clock budget \\
        7 & \texttt{functional\_correctness} & \texttt{logic\_error} & Output values outside tolerance \\
        8 & \texttt{passed} & \texttt{passed} & All checks passed \\
        \bottomrule
    \end{tabularx}
    \caption{Eight-category error taxonomy with five-bucket collapse.}
    \label{tab:taxonomy-app}
\end{table}

\paragraph{Classifier mechanism.} The deterministic classifier of \S\ref{sec:benchmark:apparatus} dispatches each iteration's build, runtime, and test logs against a fixed regex set; the first matching pattern in priority order assigns the iteration's category. Build-stage priority is \texttt{environment\_dependency} $>$ \texttt{integration} $>$ \texttt{buildability} $>$ \texttt{out\_of\_memory} $>$ \texttt{timeout}; runtime-stage priority additionally folds in \texttt{illegal\_memory\_access} and \texttt{functional\_correctness}. The auxiliary \texttt{compute-sanitizer} stream, when present, can override a runtime-stage \texttt{functional\_correctness} verdict to \texttt{illegal\_memory\_access}. Stages are evaluated in order and the first matching stage assigns the category. Within a stage, patterns are evaluated in priority order; if no pattern matches in a stage that has fired (e.g., a compile failure with unrecognized stderr), the iteration falls through to \texttt{buildability} as the compile-time default and \texttt{functional\_correctness} as the runtime default. If multiple native categories match within one stage, the iteration is flagged \texttt{unclassified} rather than coerced into a single category. Empirically, the \texttt{unclassified} flag fires on 0 of 3{,}970 iterations across the full $7 \text{ fixers} \times \numtasks{} \text{ tasks} \times K{=}5$ evaluation, so the deterministic dispatcher with explicit fallbacks achieves complete category coverage on this corpus. \Cref{tab:classifier-patterns} lists representative trigger patterns per category; the full regex set is included in the released harness.

\begin{table}[h]
    \centering
    \small
    \setlength{\tabcolsep}{4pt}
    \renewcommand{\arraystretch}{1.05}
    \begin{tabularx}{\linewidth}{>{\centering\arraybackslash}p{0.4cm} >{\raggedright\arraybackslash}p{4.5cm} >{\raggedright\arraybackslash}p{2.0cm} >{\raggedright\arraybackslash}X}
        \toprule
        \# & Category (8-cat) & Log source & Representative trigger patterns \\
        \midrule
        1 & \texttt{environment\_dependency} & nvcc / runtime stderr & \texttt{CUDA driver version is insufficient}; \texttt{unsupported gpu architecture}; \texttt{no kernel image is available for execution}; \texttt{undefined symbol} (5 sub-rules) \\
        2 & \texttt{integration}              & pre-flight / harness  & \texttt{solution.cu: No such file}; \texttt{required source references}; \texttt{undefined reference to main}; signature/I/O contract violations (5 sub-rules) \\
        3 & \texttt{buildability}             & nvcc stderr           & \texttt{syntax error}; \texttt{no matching function}; \texttt{was not declared in this scope}; \texttt{calling a \_\_host\_\_ from \_\_global\_\_}; \texttt{ptxas error} (8 sub-rules) \\
        4 & \texttt{out\_of\_memory}          & runtime stderr        & \texttt{cudaErrorMemoryAllocation}; \texttt{out of memory}; \texttt{CUDA out of memory} (1 sub-rule) \\
        5 & \texttt{illegal\_memory\_access}  & runtime / sanitizer   & \texttt{cudaErrorIllegalAddress}; \texttt{illegal memory access}; \texttt{SIGSEGV}; \texttt{invalid \_\_global\_\_ read/write} (1 sub-rule) \\
        6 & \texttt{timeout}                  & wall-clock watchdog   & \texttt{timed out}; \texttt{watchdog}; \texttt{BUILD/TEST TIMEOUT} (1 sub-rule) \\
        7 & \texttt{functional\_correctness}  & test verifier         & \texttt{tolerance/epsilon}; \texttt{shape mismatch}; \texttt{off-by-one}; \texttt{outputs differ}; \texttt{wrong result} (6 sub-rules) \\
        8 & \texttt{passed}                   & all checks            & (no patterns; harness exits zero on a clean run) \\
        \bottomrule
    \end{tabularx}
    \caption{Representative regex trigger patterns for the deterministic 8-category classifier. The full pattern set ($\sim$30 sub-rules) is shipped with the harness; each category line above gives a one-line digest of the patterns it dispatches against.}
    \label{tab:classifier-patterns}
\end{table}

\section{Stagnation Empirics}
\label{app:stagnation-thresholds}

\subsection{Per-Model Breakdown}

\Cref{tab:stagnation-app} reports the per-model stagnation breakdown on the failing portion of the corpus, by signal type. The headline trends discussed in \cref{sec:eval:stagnation} are: overall stagnation rate negatively correlates with pass@5, and the signal mix (which of the four signals dominates) is model-specific.

\begin{table}[h]
    \centering
    \small
    \setlength{\tabcolsep}{6pt}
    \renewcommand{\arraystretch}{1.05}
    \begin{tabular}{lcccccc}
        \toprule
        Model & $n_{\text{fail}}$ & overall & dup.\_code & code\_cycle & cat.\_oscill. & no\_progress \\
        \midrule
        GPT-5.4      &  89 & 85.4\% & \phantom{0}0.0\% & \phantom{0}1.1\% & 12.4\% & 71.9\% \\
        Qwen3.6-Plus & 134 & 98.5\% & 11.2\% & \phantom{0}0.7\% & \phantom{0}2.2\% & 84.3\% \\
        Gemma-4-31B-it  & 123 & 98.4\% & \phantom{0}8.1\% & \phantom{0}0.8\% & \phantom{0}4.9\% & 84.6\% \\
        Qwen3.6-27B  & 159 & 99.4\% & 49.7\% & \phantom{0}1.9\% & \phantom{0}3.1\% & 44.7\% \\
        Kimi-k2.6    & 156 & 99.4\% & 38.5\% & \phantom{0}3.8\% & \phantom{0}0.0\% & 57.1\% \\
        GLM-4.7      & 177 & 99.4\% & 50.8\% & \phantom{0}3.4\% & \phantom{0}0.6\% & 44.6\% \\
        \bottomrule
    \end{tabular}
    \caption{Per-model stagnation breakdown on failing tasks under the default protocol. $n_{\text{fail}}$ is the number of tasks that never passed within $K=5$ iterations. \emph{overall} is the fraction of those tasks where any of the four stagnation signals fired (remaining tasks reach \texttt{max\_iterations} without firing a signal); the four signal columns sum to \emph{overall} by construction.}
    \label{tab:stagnation-app}
\end{table}

\subsection{Threshold Sensitivity}

The two threshold-bearing stagnation signals introduced in \cref{sec:benchmark:metrics} are \emph{category\_oscillation}, detected when at least $m$ error-category transitions occur within the last $n$ iterations (default: 3-of-5), and \emph{no\_progress}, detected when the (category, primary error signature) tuple is unchanged across $c$ consecutive iterations (default: $c{=}3$). The other two signals (\emph{duplicate\_code}, \emph{code\_cycle}) are SHA-256-equality based and have no tunable threshold. To verify that the defaults are not load-bearing, we re-run the stagnation classifier on the stored per-iteration manifest under three variants for each signal: $\{$2-of-4 (tighter), 3-of-5 (default), 4-of-6 (looser)$\}$ for category\_oscillation and $\{c{=}2,\,3,\,4\}$ for no\_progress. This is a robustness check, not a parameter sweep; the per-model rates in \cref{tab:stagnation-app} use the defaults.

\begin{table}[h]
  \centering
  \small
  \setlength{\tabcolsep}{4pt}
  \renewcommand{\arraystretch}{1.05}
  \begin{tabular}{lcccccc}
    \toprule
    & \multicolumn{3}{c}{cat\_oscill ($m$-of-$n$)} & \multicolumn{3}{c}{no\_progress ($c$-cons.)} \\
    \cmidrule(lr){2-4}\cmidrule(lr){5-7}
    Model & 2-of-4 & \textbf{3-of-5} & 4-of-6 & $c{=}2$ & $\boldsymbol{c{=}3}$ & $c{=}4$ \\
    \midrule
    GLM-4.7        & 100.0\% & 99.4\% & 98.8\% & 100.0\% & 99.4\% & 54.8\% \\
    Kimi-k2.6      & 100.0\% & 99.4\% & 99.4\% & 100.0\% & 99.4\% & 42.3\% \\
    Qwen3.6-27B    & 100.0\% & 99.4\% & 96.2\% & 100.0\% & 99.4\% & 54.7\% \\
    Qwen3.6-Plus   & 100.0\% & 98.5\% & 96.3\% & 100.0\% & 98.5\% & 14.2\% \\
    Gemma-4-31B-it    & 100.0\% & 98.4\% & 93.5\% & 100.0\% & 98.4\% & 13.8\% \\
    GPT-5.4        & 100.0\% & 85.4\% & 73.0\% & 100.0\% & 85.4\% & 13.5\% \\
    \bottomrule
  \end{tabular}
  \caption{Overall stagnation rate (\%) under threshold variants for \emph{category\_oscillation} and \emph{no\_progress}; the bold column in each block is the default reported in \cref{tab:stagnation-app}. Tighter thresholds (2-of-4, $c{=}2$) saturate at 100\% on every fixer because every failing task contains at least one category transition or one repeated (category, error) tuple. Looser \emph{category\_oscillation} (4-of-6) shifts modestly ($\leq 12.4$ percentage points), but looser \emph{no\_progress} ($c{=}4$) collapses by 30--86 percentage points because four consecutive identical (category, error) iterations cannot be observed within the $K{=}5$ iteration budget for many trajectories. The default thresholds (3-of-5, $c{=}3$) are therefore not arbitrary but \emph{calibrated to $K{=}5$}: tighter saturates, looser starves. Model rank order is preserved across all six variants.}
  \label{tab:stagnation-thresholds}
\end{table}

\section{Reproducibility}
\label{app:reproducibility}

\paragraph{Hardware.} Pilot runs use a single NVIDIA RTX~PRO~6000 Blackwell Server Edition (compute capability 12.0). Full-corpus runs reported in the camera-ready will add one Hopper-class GPU (H200, sm\_90) and one Ampere-class GPU (A100, sm\_80), which together with the Blackwell host cover the three architecture buckets under which the corpus is stratified.

\paragraph{Software.} CUDA 12.4, Python 3.11, PyTorch 2.4.0. Evaluation framework in $\sim$4{,}000 lines of Python across 13 modules, with a 91-case unit-test suite. Reproducibility artifacts (frozen config snapshot, per-iteration manifest, raw model responses, deduplicated solution code) are written to a timestamped result directory and made available with the camera-ready release.

\paragraph{Kernel timing protocol.}\label{app:timing} Each candidate kernel binary is re-launched 10 times post hoc; a per-GPU \texttt{flock} serializes launches so concurrent debugger workers do not collide on the CUDA driver or \texttt{nsys} capture lock. Within each launch, the upstream harness executes its own timed-trial loop, following each task's source-pool convention: 100 cuda-event timed trials with three warmup iterations and an L2 cache flush per trial for KernelBench-pool tasks, ten timed trials for CUDABench/CUTLASS-pool tasks, and self-reported wall-clock for ComputeEval-pool tasks. Speedup is computed as $\mathrm{reference\_mean\_ms} / \mathrm{candidate\_mean\_ms}$. The coefficient of variation across the ten outer means, $CV=\sigma/\mu$, is reported alongside each speedup as a quality flag at the 3\% threshold; the threshold is reported, not enforced, so high-CV cells are retained to avoid selection bias. Across 196 measured cells, the CV distribution has median 0.8\% and p90 9.3\%; 52/196 (26.5\%) exceed 3\% and are retained. A small fraction of passed cells, 48/232 ($\approx 21\%$), fall back to single-launch timing because the candidate is \texttt{incompatible\_arch} for the available GPUs or fails the perf-mode build; these cells use the harness single-shot mean and omit CV.

\paragraph{Release artifacts.}
\begin{itemize}[leftmargin=1.5em]
    \item \textbf{Corpus.} Full \numtasks-task manifest with source provenance per task.
    \item \textbf{Harness.} Dispatcher, three execution backends, pre-flight checker, classifier, feedback generator, stagnation detector.
    \item \textbf{Analyzer.} Metric pipeline reproducing the \nummetrics-layer metric suite from stored manifests.
    \item \textbf{Configs.} YAML configs for each model/budget combination reported in the paper.
    \item \textbf{Logs.} Per-iteration manifest, raw build logs, raw model responses, deduplicated solution code.
\end{itemize}

\section{Per-axis Raw Numbers and Metric Breakdowns}
\label{app:per-axis-raw}

\begin{table}[H]
    \centering
    \footnotesize
    \setlength{\tabcolsep}{4pt}
    \renewcommand{\arraystretch}{0.95}
    \begin{tabular}{lcccccccc}
        \toprule
        Model & $p{=}0.0$ & $0.1$ & $0.3$ & $0.5$ & \textbf{0.7} & $0.9$ & $1.0$ & $1.5$ \\
        \midrule
        GPT-5.4       & 55.5\% & 37.5\% & 37.0\% & 32.0\% & \textbf{30.0\%} & 27.0\% & 24.5\% & 15.5\% \\
        Qwen3.6-Plus  & 33.0\% & 31.0\% & 29.0\% & 26.0\% & \textbf{22.0\%} & 19.0\% & 15.0\% & 8.0\%  \\
        Gemma-4-31B-it   & 26.5\% & 26.0\% & 25.0\% & 22.5\% & \textbf{20.0\%} & 17.0\% & 13.5\% & 8.0\%  \\
        Qwen3.6-27B   & 24.5\% & 23.5\% & 23.0\% & 22.5\% & \textbf{20.5\%} & 18.5\% & 16.5\% & 9.5\%  \\
        Kimi-k2.6     & 22.0\% & 21.0\% & 20.5\% & 20.0\% & \textbf{19.5\%} & 19.0\% & 15.0\% & 7.0\%  \\
        GLM-4.7       & 15.5\% & 15.5\% & 15.5\% & 15.5\% & \textbf{13.5\%} & 12.5\% & 12.0\% & 6.5\%  \\
        MiniMax-M2.7  & 13.5\% & 13.0\% & 12.5\% & 11.5\% & \textbf{11.0\%} & 9.5\%  & 8.5\%  & 5.0\%  \\
        \bottomrule
    \end{tabular}
    \caption{$A_1$: pass@5 vs.\ performance-gate threshold $p$ (full sweep). Bold = paper-headline default $p=0.7$.}
    \label{tab:axis-p-app}
\end{table}
\vspace{-1.2em}

\begin{table}[H]
    \centering
    \footnotesize
    \setlength{\tabcolsep}{6pt}
    \renewcommand{\arraystretch}{0.95}
    \begin{tabular}{lccc}
        \toprule
        Model & Iter & Rep & $\Delta$ \\
        \midrule
        GPT-5.4       & 29.1\% & 34.7\% & $-5.6$ \\
        Qwen3.6-Plus  & 23.0\% & 13.1\% & $\phantom{+}+9.9$ \\
        Gemma-4-31B-it   & 21.6\% & 22.1\% & $-0.5$ \\
        Qwen3.6-27B   & 19.2\% & 13.6\% & $\phantom{+}+5.6$ \\
        Kimi-k2.6     & 18.8\% & 21.1\% & $-2.3$ \\
        GLM-4.7       & 12.7\% & 12.2\% & $\phantom{+}+0.5$ \\
        MiniMax-M2.7  & 10.3\% & 8.9\%  & $\phantom{+}+1.4$ \\
        \bottomrule
    \end{tabular}
    \caption{$A_2$: Iter ($T{=}0.7$) vs.\ Rep ($T{=}1.0$) at the same 5-call budget.}
    \label{tab:axis-method-app}
\end{table}
\vspace{-1.2em}

\begin{table}[H]
    \centering
    \footnotesize
    \setlength{\tabcolsep}{8pt}
    \renewcommand{\arraystretch}{0.95}
    \begin{tabular}{lccccc}
        \toprule
        Model        & $L_0$ & $L_1$ & $L_2$ & $L_3$ & $L_3^{*}$ \\
        \midrule
        GPT-5.4      & 31.0\% & 30.5\% & 41.3\%          & 29.1\%          & \textbf{43.7\%} \\
        Qwen3.6-Plus & 16.9\% & 18.3\% & 24.4\%          & 23.0\%          & \textbf{25.8\%} \\
        Kimi-k2.6    & 14.6\% & 14.1\% & \textbf{23.9\%} & 18.8\%          & 20.7\% \\
        Gemma-4-31B-it  & 18.3\% & 16.4\% & 19.7\%          & 21.6\%          & \textbf{22.1\%} \\
        Qwen3.6-27B  & \phantom{0}9.9\%  & \phantom{0}9.9\%  & 13.1\%   & \textbf{19.2\%} & 15.0\% \\
        GLM-4.7      & 11.7\% & 12.7\% & 15.5\%          & \textbf{16.9\%} & 14.1\% \\
        \bottomrule
    \end{tabular}
    \caption{$A_3$: pass@5 across feedback levels. Bold = per-model best. $L_3^{*}$ = $L_3$-raw (raw \texttt{stderr} without category-aware reformatting).}
    \label{tab:axis-feedback-app}
\end{table}
\vspace{-1.2em}

\begin{table}[H]
    \centering
    \footnotesize
    \setlength{\tabcolsep}{8pt}
    \renewcommand{\arraystretch}{0.95}
    \begin{tabular}{lcccc}
        \toprule
        Model        & $H_1$ & $H_2$ & $H_3$ & $H_4$ \\
        \midrule
        GPT-5.4      & \textbf{33.3\%} & 30.5\% & 30.0\% & 29.1\% \\
        Gemma-4-31B-it  & \textbf{27.7\%} & 21.6\% & 19.2\% & 21.6\% \\
        Kimi-k2.6    & 24.4\% & 21.6\% & \textbf{26.3\%} & 18.8\% \\
        Qwen3.6-Plus & 19.7\% & \textbf{23.9\%} & \textbf{23.9\%} & 23.0\% \\
        Qwen3.6-27B  & 18.3\% & 16.0\% & 15.5\% & \textbf{19.2\%} \\
        GLM-4.7      & \textbf{16.9\%} & 16.4\% & 16.4\% & 12.7\% \\
        \bottomrule
    \end{tabular}
    \caption{$A_4$: pass@5 across $H_1$--$H_4$ (default $H_4$). Bold = best per model.}
    \label{tab:axis-H-app}
\end{table}
\vspace{-1.2em}

\begin{table}[H]
    \centering
    \footnotesize
    \setlength{\tabcolsep}{6pt}
    \renewcommand{\arraystretch}{0.95}
    \begin{tabular}{lccccc}
        \toprule
        Model & pass@1 & \texttt{debug\_rate@5} & pass@5 & stagnation\% & dominant signal \\
        \midrule
        GPT-5.4       & 14.5\% & 17.1\% & 29.1\% & 85.4\% & \texttt{no\_progress}    \\
        Qwen3.6-Plus  & 15.5\% & \phantom{0}8.9\%  & 23.0\% & 98.5\% & \texttt{no\_progress}    \\
        Gemma-4-31B-it   & 14.7\% & \phantom{0}8.1\%  & 21.6\% & 98.4\% & \texttt{no\_progress}    \\
        Qwen3.6-27B   & 10.3\% & \phantom{0}9.9\%  & 19.2\% & 99.4\% & \texttt{duplicate\_code} \\
        Kimi-k2.6     & 13.6\% & \phantom{0}6.0\%  & 18.8\% & 99.4\% & \texttt{no\_progress}    \\
        GLM-4.7       & \phantom{0}7.5\%  & \phantom{0}5.6\%  & 12.7\% & 99.4\% & \texttt{duplicate\_code} \\
        \bottomrule
    \end{tabular}
    \caption{Metric family snapshot per fixer under the default protocol ($K=5$, $H=4$, feedback $L_3$, iterative $T{=}0.7$, performance-gate $p{=}0.7$). \texttt{debug\_rate@5} decomposes pass@5; stagnation\% is the fraction of \emph{failing} tasks where one of the four signals fired; dominant signal is the most-frequent. Sorted by pass@5 descending.}
    \label{tab:metric-snapshot-app}
\end{table}
\vspace{-1.2em}

\begin{table}[H]
    \centering
    \footnotesize
    \setlength{\tabcolsep}{4pt}
    \renewcommand{\arraystretch}{0.95}
    \begin{tabular}{lccccc}
        \toprule
        Model & \texttt{compile} & \texttt{logic} & \texttt{perf-broken} & \texttt{mem-crash} & \texttt{timeout} \\
              & ($n{=}101$)      & ($n{=}40$)     & ($n{=}41$)           & ($n{=}16$)         & ($n{=}2$)        \\
        \midrule
        GPT-5.4       & 47.5\% & 60.0\% & 70.7\% & 56.2\% & 50.0\% \\
        Qwen3.6-Plus  & 26.7\% & 35.0\% & 43.9\% & 37.5\% & 50.0\% \\
        Gemma-4-31B-it   & 24.4\% & 29.4\% & 39.0\% & 53.8\% & 50.0\% \\
        Qwen3.6-27B   & 22.8\% & 20.0\% & 31.7\% & 25.0\% & 50.0\% \\
        Kimi-k2.6     & 20.8\% & 10.0\% & 34.1\% & 25.0\% & 50.0\% \\
        GLM-4.7       & 17.8\% &  7.5\% & 24.4\% &  0.0\% &  0.0\% \\
        MiniMax-M2.7  & 12.9\% & 12.5\% & 12.2\% & 18.8\% & 50.0\% \\
        \bottomrule
    \end{tabular}
    \caption{Per-fixer \texttt{fix\_rate} per 5-bucket curation-time error category (\S\ref{sec:benchmark:metrics}, \cref{app:taxonomy}). Each cell is the fraction of broken-starts in that bucket that the fixer eventually solved at any iteration under correctness-only scoring ($p{=}0$). \texttt{timeout} is sparse ($n{=}2$) and not interpretable. Gemma-4-31B-it's denominators are smaller ($n{=}78/34/41/13/2$ for compile/logic/perf-broken/mem-crash/timeout) because its manifest covers 168 of 200 v2 tasks; rates are reported within available coverage.}
    \label{tab:fix-rate-by-bucket}
\end{table}

\section{Corpus Composition}

\begin{table}[H]
    \centering
    \footnotesize
    \setlength{\tabcolsep}{5pt}
    \renewcommand{\arraystretch}{0.95}
    \begin{tabular}{@{}lcl@{}}
        \toprule
        Family & Instances & Upstream attribution \\
        \midrule
        \multicolumn{3}{@{}l}{\emph{NVIDIA-library and ours-curated}} \\
        \midrule
        KernelBench                         & 84  & \citet{ouyang2025kernelbench} \\
        \texttt{CUDA}                       & 49  & Curated CUDA samples \\
        LayerNorm                           & 16  & NVIDIA Apex LayerNorm \\
        cuSOLVER                            & 10  & NVIDIA cuSOLVER \\
        cuFFT samples                       & 9   & NVIDIA cuFFT samples \\
        cuSPARSE                            & 6   & NVIDIA cuSPARSE \\
        ThunderKittens                      & 4   & \citet{thunderkittensrepo} \\
        cuBLAS                              & 3   & NVIDIA cuBLAS \\
        cuFFT                               & 2   & NVIDIA cuFFT \\
        \midrule
        \multicolumn{3}{@{}l}{\emph{Frontier-kernel}} \\
        \midrule
        CUTLASS                             & 20  & NVIDIA \citep{cutlassrepo} \\
        FlashAttention v2                   & 10  & \cite{fa2} \\
        \midrule
        Total                               & \numtasks{} & \\
        \bottomrule
    \end{tabular}
    \caption{Corpus instances by kernel family (11 families, two tiers).}
    \label{tab:families-app}
\end{table}
\vspace{-1.2em}

\begin{table}[H]
\centering
\footnotesize
\setlength{\tabcolsep}{5pt}
\renewcommand{\arraystretch}{0.95}
\begin{minipage}[t]{0.50\linewidth}
\centering
\begin{tabular}{lc}
    \toprule
    Source model & Instances \\
    \midrule
    Kimi-k2.5/k2.6     & 85 (39.9\%) \\
    GPT-5.4            & 46 (21.6\%) \\
    Gemma-4-31B-it        & 32 (15.0\%) \\
    Qwen3.5-122B-A10B  & 21 (9.9\%)  \\
    Manual injection   & 17 (8.0\%)  \\
    Qwen3.6-Plus       & 12 (5.6\%)  \\
    \midrule
    Total              & \numtasks{} (100\%) \\
    \bottomrule
\end{tabular}
\caption{Broken\_start source distribution. The ``Manual injection'' protocol is detailed in the paragraph following \cref{tab:source-full}.}
\label{tab:sources-app}
\end{minipage}\hfill
\begin{minipage}[t]{0.46\linewidth}
\centering
\begin{tabular}{lc}
    \toprule
    Error category & Instances \\
    \midrule
    \texttt{compile\_error}  & 105 (49.3\%) \\
    \texttt{logic\_error}    & 45 (21.1\%) \\
    \texttt{perf\_broken}    & 41 (19.2\%) \\
    \texttt{memory\_crash}   & 18 (8.5\%) \\
    \texttt{timeout}         & 4 (1.9\%) \\
    \\
    \midrule
    Total                    & \numtasks{} (100\%) \\
    \bottomrule
\end{tabular}
\caption{5-bucket error-category distribution used at curation time, collapsed from the \numcategories{}-category per-iteration taxonomy (full mapping in \cref{tab:taxonomy-app}). \texttt{perf\_broken} is broken-start-only.}
\label{tab:errors-app}
\end{minipage}
\end{table}
\vspace{-1.2em}

\begin{table}[H]
    \centering
    \scriptsize
    \setlength{\tabcolsep}{4pt}
    \renewcommand{\arraystretch}{0.95}
    \begin{tabular}{p{3.0cm} p{3.0cm} cccccc}
        \toprule
        Family & Backend & \texttt{compile} & \texttt{logic} & \texttt{mem} & \texttt{timeout} & \texttt{perf} & Total \\
        \midrule
        \multicolumn{8}{l}{\emph{NVIDIA-library and ours-curated (183 instances)}} \\
        \midrule
        KernelBench             & KB static + runtime   & 33 & 32 & 15 & 4 & 0  & 84 \\
        CUDA samples            & Raw \texttt{nvcc}     & 41 & 5  & 3  & 0 & 0  & 49 \\
        LayerNorm               & Project Makefile      & 0  & 0  & 0  & 0 & 16 & 16 \\
        cuSOLVER                & Raw \texttt{nvcc}     & 10 & 0  & 0  & 0 & 0  & 10 \\
        cuFFT samples           & Project Makefile      & 0  & 0  & 0  & 0 & 9  & 9 \\
        cuSPARSE                & Raw \texttt{nvcc}     & 6  & 0  & 0  & 0 & 0  & 6 \\
        ThunderKittens          & Project Makefile      & 4  & 0  & 0  & 0 & 0  & 4 \\
        cuBLAS                  & Raw \texttt{nvcc}     & 3  & 0  & 0  & 0 & 0  & 3 \\
        cuFFT                   & Raw \texttt{nvcc}     & 2  & 0  & 0  & 0 & 0  & 2 \\
        \midrule
        \multicolumn{8}{l}{\emph{Frontier-kernel (30 instances)}} \\
        \midrule
        CUTLASS                 & Project Makefile      & 6  & 8  & 0  & 0 & 6  & 20 \\
        FlashAttention v2       & Project Makefile      & 0  & 0  & 0  & 0 & 10 & 10 \\
        \midrule
        \textbf{Total} & & \textbf{105} & \textbf{45} & \textbf{18} & \textbf{4} & \textbf{41} & \textbf{\numtasks{}} \\
        \bottomrule
    \end{tabular}
    \caption{Family $\times$ 5-bucket error-category cross-tabulation across the 11-family corpus.}
    \label{tab:source-full}
\end{table}

\paragraph{Manual injection.} The 17 ``Manual injection'' rows in \cref{tab:sources-app} denote curator-modified instances where the source LLM's iter-1 output passed correctness or performance-gate checks; the authors then edited the kernel to surface one of the five curation-time failure buckets and re-ran the harness to confirm deterministic reproduction of the targeted failure under the standard build/test commands. This preserves the (broken\_kernel, error\_log) reproducibility invariant of \S\ref{sec:benchmark:protocol} but relaxes the ``failure originated from the LLM'' provenance for these specific instances.

%% file: main.bbl
\begin{thebibliography}{39}
\providecommand{\natexlab}[1]{#1}
\providecommand{\url}[1]{\texttt{#1}}
\expandafter\ifx\csname urlstyle\endcsname\relax
  \providecommand{\doi}[1]{doi: #1}\else
  \providecommand{\doi}{doi: \begingroup \urlstyle{rm}\Url}\fi

\bibitem[Alzahrani et~al.(2024)Alzahrani, Alyahya, Alnumay, Alrashed, Alsubaie,
  Almushayqih, Mirza, Alotaibi, Al-Twairesh, Alowisheq,
  et~al.]{alzahrani2024benchmarks}
Norah Alzahrani, Hisham Alyahya, Yazeed Alnumay, Sultan Alrashed, Shaykhah
  Alsubaie, Yousef Almushayqih, Faisal Mirza, Nouf Alotaibi, Nora Al-Twairesh,
  Areeb Alowisheq, et~al.
\newblock When benchmarks are targets: Revealing the sensitivity of large
  language model leaderboards.
\newblock In \emph{Proceedings of the 62nd Annual Meeting of the Association
  for Computational Linguistics (Volume 1: Long Papers)}, pages 13787--13805,
  2024.

\bibitem[Baronio et~al.(2025)Baronio, Marsella, Pan, Guo, and
  Alberti]{baronio2025kevin}
Carlo Baronio, Pietro Marsella, Ben Pan, Simon Guo, and Silas Alberti.
\newblock Kevin: Multi-turn rl for generating cuda kernels.
\newblock \emph{arXiv preprint arXiv:2507.11948}, 2025.

\bibitem[Brent(1980)]{brent1980improved}
Richard~P Brent.
\newblock An improved monte carlo factorization algorithm.
\newblock \emph{BIT Numerical Mathematics}, 20\penalty0 (2):\penalty0 176--184,
  1980.

\bibitem[Chen et~al.(2021)Chen, Tworek, Jun, Yuan, Pinto, Kaplan, Edwards,
  Burda, Joseph, Brockman, et~al.]{chen2021evaluating}
Mark Chen, Jerry Tworek, Heewoo Jun, Qiming Yuan, Henrique Ponde De~Oliveira
  Pinto, Jared Kaplan, Harri Edwards, Yuri Burda, Nicholas Joseph, Greg
  Brockman, et~al.
\newblock Evaluating large language models trained on code.
\newblock \emph{arXiv preprint arXiv:2107.03374}, 2021.

\bibitem[Chen et~al.(2026)Chen, Ye, Xu, Ye, Liu, Hassani, Chen, Kerr, Wu, Xu,
  et~al.]{avo}
Terry Chen, Zhifan Ye, Bing Xu, Zihao Ye, Timmy Liu, Ali Hassani, Tianqi Chen,
  Andrew Kerr, Haicheng Wu, Yang Xu, et~al.
\newblock Avo: Agentic variation operators for autonomous evolutionary search.
\newblock \emph{arXiv preprint arXiv:2603.24517}, 2026.

\bibitem[Dao(2023)]{fa2}
Tri Dao.
\newblock Flashattention-2: Faster attention with better parallelism and work
  partitioning, 2023.
\newblock URL \url{https://arxiv.org/abs/2307.08691}.

\bibitem[De~Jong(1975)]{de1975analysis}
Kenneth~Alan De~Jong.
\newblock \emph{An analysis of the behavior of a class of genetic adaptive
  systems.}
\newblock University of Michigan, 1975.

\bibitem[Flajolet and Odlyzko(1989)]{flajolet1989random}
Philippe Flajolet and Andrew~M Odlyzko.
\newblock Random mapping statistics.
\newblock In \emph{Workshop on the Theory and Application of of Cryptographic
  Techniques}, pages 329--354. Springer, 1989.

\bibitem[Hariri et~al.(2025)Hariri, Samandar, Hinczewski, and
  Chaudhary]{hariri2025don}
Mohsen Hariri, Amirhossein Samandar, Michael Hinczewski, and Vipin Chaudhary.
\newblock Don't pass@ k: A bayesian framework for large language model
  evaluation.
\newblock \emph{arXiv preprint arXiv:2510.04265}, 2025.

\bibitem[Holtzman et~al.(2019)Holtzman, Buys, Du, Forbes, and
  Choi]{holtzman2019curious}
Ari Holtzman, Jan Buys, Li~Du, Maxwell Forbes, and Yejin Choi.
\newblock The curious case of neural text degeneration.
\newblock \emph{arXiv preprint arXiv:1904.09751}, 2019.

\bibitem[Jimenez et~al.(2023)Jimenez, Yang, Wettig, Yao, Pei, Press, and
  Narasimhan]{jimenez2023swe}
Carlos~E Jimenez, John Yang, Alexander Wettig, Shunyu Yao, Kexin Pei, Ofir
  Press, and Karthik~R Narasimhan.
\newblock Swe-bench: Can language models resolve real-world github issues?
\newblock In \emph{The twelfth international conference on learning
  representations}, 2023.

\bibitem[Knuth(2014)]{knuth2014art}
Donald~E Knuth.
\newblock \emph{The art of computer programming: Seminumerical algorithms,
  volume 2}.
\newblock Addison-Wesley Professional, 2014.

\bibitem[Li et~al.(2025)Li, Li, Gao, Shi, Li, Wang, Huang, WangHaojie, Wang,
  Han, et~al.]{li2025tritonbench}
Jianling Li, Shangzhan Li, Zhenye Gao, Qi~Shi, Yuxuan Li, Zefan Wang, Jiacheng
  Huang, WangHaojie WangHaojie, Jianrong Wang, Xu~Han, et~al.
\newblock Tritonbench: Benchmarking large language model capabilities for
  generating triton operators.
\newblock In \emph{Findings of the Association for Computational Linguistics:
  ACL 2025}, pages 23053--23066, 2025.

\bibitem[Li et~al.(2026)Li, Zhang, Chen, Luo, Hong, and Ding]{stitchcuda}
Shiyang Li, Zijian Zhang, Winson Chen, Yuebo Luo, Mingyi Hong, and Caiwen Ding.
\newblock Stitchcuda: An automated multi-agents end-to-end gpu programing
  framework with rubric-based agentic reinforcement learning, 2026.

\bibitem[Liu et~al.(2026)Liu, Xu, Li, Zheng, Li, Liu, and He]{liu2026dr}
Wei Liu, Jiawei Xu, Yingru Li, Longtao Zheng, Tianjian Li, Qian Liu, and
  Junxian He.
\newblock Dr. kernel: Reinforcement learning done right for triton kernel
  generations.
\newblock \emph{arXiv preprint arXiv:2602.05885}, 2026.

\bibitem[Madaan et~al.(2023)Madaan, Tandon, Gupta, Hallinan, Gao, Wiegreffe,
  Alon, Dziri, Prabhumoye, Yang, Gupta, Majumder, Hermann, Welleck,
  Yazdanbakhsh, and Clark]{madaan2023selfrefineiterativerefinementselffeedback}
Aman Madaan, Niket Tandon, Prakhar Gupta, Skyler Hallinan, Luyu Gao, Sarah
  Wiegreffe, Uri Alon, Nouha Dziri, Shrimai Prabhumoye, Yiming Yang, Shashank
  Gupta, Bodhisattwa~Prasad Majumder, Katherine Hermann, Sean Welleck, Amir
  Yazdanbakhsh, and Peter Clark.
\newblock Self-refine: Iterative refinement with self-feedback, 2023.
\newblock URL \url{https://arxiv.org/abs/2303.17651}.

\bibitem[MiniMax et~al.(2025)MiniMax, :, Chen, Li, Gong, Jiang, Fei, Yang,
  Shan, Yu, Wang, Zhu, Xiao, Du, Zhang, Qiao, Zhang, Du, Guo, Chen, Ding, Sun,
  Li, Jiao, Zhou, Zhang, Ding, Sun, Feng, Cai, Zhu, Sun, Zhuang, Cai, Song,
  Zhu, Li, Tian, Liu, Xu, Yan, Liu, He, Feng, Yang, Xiao, Han, Wang, Yu, Feng,
  Li, Zheng, Du, Yang, Zeng, Yu, Tao, Chi, Zhang, Lin, Hu, Di, Gao, Li, Zhao,
  Ren, Xu, Li, Wang, Tian, Leng, Chen, Chen, Shi, Weng, Guan, Yu, Li, Zhu, Li,
  Cai, Liang, Cheng, Kong, Li, Chen, Song, Luo, Su, Li, Han, Hou, Lu, Zou,
  Shen, Gong, Ma, Wang, Shi, Zhong, Duan, Fu, Hu, Gao, Fan, Yang, Li, Hu,
  Huang, Li, Xu, Mao, Shi, Wenren, Li, Li, Tian, Zhu, Fan, Wu, Xu, Yu, Lyu,
  Jiang, Gao, Wu, Song, and Sun]{minimax2025minimaxm1scalingtesttimecompute}
MiniMax, :, Aili Chen, Aonian Li, Bangwei Gong, Binyang Jiang, Bo~Fei, Bo~Yang,
  Boji Shan, Changqing Yu, Chao Wang, Cheng Zhu, Chengjun Xiao, Chengyu Du, Chi
  Zhang, Chu Qiao, Chunhao Zhang, Chunhui Du, Congchao Guo, Da~Chen, Deming
  Ding, Dianjun Sun, Dong Li, Enwei Jiao, Haigang Zhou, Haimo Zhang, Han Ding,
  Haohai Sun, Haoyu Feng, Huaiguang Cai, Haichao Zhu, Jian Sun, Jiaqi Zhuang,
  Jiaren Cai, Jiayuan Song, Jin Zhu, Jingyang Li, Jinhao Tian, Jinli Liu,
  Junhao Xu, Junjie Yan, Junteng Liu, Junxian He, Kaiyi Feng, Ke~Yang, Kecheng
  Xiao, Le~Han, Leyang Wang, Lianfei Yu, Liheng Feng, Lin Li, Lin Zheng, Linge
  Du, Lingyu Yang, Lunbin Zeng, Minghui Yu, Mingliang Tao, Mingyuan Chi, Mozhi
  Zhang, Mujie Lin, Nan Hu, Nongyu Di, Peng Gao, Pengfei Li, Pengyu Zhao,
  Qibing Ren, Qidi Xu, Qile Li, Qin Wang, Rong Tian, Ruitao Leng, Shaoxiang
  Chen, Shaoyu Chen, Shengmin Shi, Shitong Weng, Shuchang Guan, Shuqi Yu,
  Sichen Li, Songquan Zhu, Tengfei Li, Tianchi Cai, Tianrun Liang, Weiyu Cheng,
  Weize Kong, Wenkai Li, Xiancai Chen, Xiangjun Song, Xiao Luo, Xiao Su, Xiaobo
  Li, Xiaodong Han, Xinzhu Hou, Xuan Lu, Xun Zou, Xuyang Shen, Yan Gong, Yan
  Ma, Yang Wang, Yiqi Shi, Yiran Zhong, Yonghong Duan, Yongxiang Fu, Yongyi Hu,
  Yu~Gao, Yuanxiang Fan, Yufeng Yang, Yuhao Li, Yulin Hu, Yunan Huang, Yunji
  Li, Yunzhi Xu, Yuxin Mao, Yuxuan Shi, Yuze Wenren, Zehan Li, Zelin Li, Zhanxu
  Tian, Zhengmao Zhu, Zhenhua Fan, Zhenzhen Wu, Zhichao Xu, Zhihang Yu, Zhiheng
  Lyu, Zhuo Jiang, Zibo Gao, Zijia Wu, Zijian Song, and Zijun Sun.
\newblock Minimax-m1: Scaling test-time compute efficiently with lightning
  attention, 2025.
\newblock URL \url{https://arxiv.org/abs/2506.13585}.

\bibitem[Mizrahi et~al.(2024)Mizrahi, Kaplan, Malkin, Dror, Shahaf, and
  Stanovsky]{mizrahi2024state}
Moran Mizrahi, Guy Kaplan, Dan Malkin, Rotem Dror, Dafna Shahaf, and Gabriel
  Stanovsky.
\newblock State of what art? a call for multi-prompt llm evaluation.
\newblock \emph{Transactions of the Association for Computational Linguistics},
  12:\penalty0 933--949, 2024.

\bibitem[{NVIDIA}(2025)]{computeeval2025}
{NVIDIA}.
\newblock Computeeval: Evaluating large language models for cuda code
  generation.
\newblock GitHub repository, 2025.
\newblock URL \url{https://github.com/NVIDIA/compute-eval}.

\bibitem[{NVIDIA}(2026)]{cutlassrepo}
{NVIDIA}.
\newblock Cutlass.
\newblock GitHub repository, 2026.
\newblock URL \url{https://github.com/NVIDIA/cutlass}.

\bibitem[Olausson et~al.(2023)Olausson, Inala, Wang, Gao, and
  Solar-Lezama]{olausson2023self}
Theo~X Olausson, Jeevana~Priya Inala, Chenglong Wang, Jianfeng Gao, and Armando
  Solar-Lezama.
\newblock Is self-repair a silver bullet for code generation?
\newblock \emph{arXiv preprint arXiv:2306.09896}, 2023.

\bibitem[Ouyang et~al.(2025)Ouyang, Guo, Arora, Zhang, Hu, R{\'e}, and
  Mirhoseini]{ouyang2025kernelbench}
Anne Ouyang, Simon Guo, Simran Arora, Alex~L Zhang, William Hu, Christopher
  R{\'e}, and Azalia Mirhoseini.
\newblock Kernelbench: Can llms write efficient gpu kernels?
\newblock \emph{arXiv preprint arXiv:2502.10517}, 2025.

\bibitem[Peng et~al.(2025)Peng, Gotmare, Lyu, Xiong, Savarese, and
  Sahoo]{peng2025perfcodegen}
Yun Peng, Akhilesh~Deepak Gotmare, Michael~R Lyu, Caiming Xiong, Silvio
  Savarese, and Doyen Sahoo.
\newblock Perfcodegen: Improving performance of llm generated code with
  execution feedback.
\newblock In \emph{2025 IEEE/ACM Second International Conference on AI
  Foundation Models and Software Engineering (Forge)}, pages 1--13. IEEE, 2025.

\bibitem[Saltelli et~al.(2008)Saltelli, Ratto, Andres, Campolongo, Cariboni,
  Gatelli, Saisana, and Tarantola]{saltelli2008global}
Andrea Saltelli, Marco Ratto, Terry Andres, Francesca Campolongo, Jessica
  Cariboni, Debora Gatelli, Michaela Saisana, and Stefano Tarantola.
\newblock \emph{Global sensitivity analysis: the primer}.
\newblock John Wiley \& Sons, 2008.

\bibitem[Sclar et~al.(2023)Sclar, Choi, Tsvetkov, and
  Suhr]{sclar2023quantifying}
Melanie Sclar, Yejin Choi, Yulia Tsvetkov, and Alane Suhr.
\newblock Quantifying language models' sensitivity to spurious features in
  prompt design or: How i learned to start worrying about prompt formatting.
\newblock \emph{arXiv preprint arXiv:2310.11324}, 2023.

\bibitem[Singh et~al.(2026)Singh, Fry, Perelman, Tart, Ganesh, El-Kishky,
  McLaughlin, Low, Ostrow, Ananthram, Nathan, Luo, Helyar, Madry, Efremov,
  Spyra, Baker-Whitcomb, Beutel, Karpenko, Makelov, Neitz, Wei, Barr,
  Kirchmeyer, Ivanov, Christakis, Gillespie, Tam, Bennett, Wan, Huang,
  Sandjideh, Yang, Kumar, Saraiva, Vallone, Gheorghe, Garcia, Braunstein, Liu,
  Schmidt, Mereskin, Mishchenko, Applebaum, Rogerson, Rajan, Wei, Kotha,
  Srivastava, Agrawal, Vijayvergiya, Tyra, Nair, Nayak, Eggers, Ji, Hoover,
  Chen, Chen, Barak, Minaiev, Hao, Baker, Lightcap, McKinzie, Wang, Quinn,
  Fioca, Hsu, Yang, Yu, Zhang, Brenner, Zetino, Raymond, Lugaresi, Paz, Hudson,
  Whitney, Li, Chen, Cole, Voss, Ding, Shen, Huang, Colby, Hallacy, Koch, Lu,
  Kaplan, Kim, Minott-Henriques, Frey, Yu, Czarnecki, Reid, Wei, Decareaux,
  Scheau, Zhang, Forbes, Tang, Goldberg, Roberts, Palmie, Kappler, Levine,
  Wright, Leo, Lin, Robinson, Grabb, Chen, Lim, Salama, Bhattacharjee, Tsipras,
  Li, Yu, Strouse, Williams, Hunn, Bayes, Arbus, Akyurek, Le, Widmann, Yani,
  Proehl, Sert, Cheung, Schwartz, Han, Jiang, Mitchell, Sigler, Wallace,
  Ritter, Kavanaugh, Mays, Nikishin, Li, Such, de~Avila Belbute~Peres, Raso,
  Bekerman, Tsimpourlas, Chantzis, Song, Zhang, Raila, McGrath, Briggs, Yang,
  Parascandolo, Chabot, Kim, Zhao, Valiant, Leclerc, Salman, Wang, Sheng,
  Jiang, Wang, Jin, Sikchi, Schmidt, Aspegren, Chen, Qiu, Lightman, Covert,
  Kivlichan, Silber, Sohl, Hammoud, Clavera, Lan, Akkaya, Kostrikov, Kofman,
  Etinger, Singal, Hehir, Huh, Pan, Wilczynski, Pachocki, Lee, Quinn, Kiros,
  Kalra, Samaroo, Wang, Wolfe, Chen, Wang, Harb, Han, Wang, Zhao, Chen, Yang,
  Tworek, Chand, Landon, Liang, Lin, Liu, Wang, Tang, Yin, Jang, Morris, Flynn,
  Ferstad, Heidecke, Fishbein, Hallman, Grant, Chien, Gordon, Park, Liss,
  Kraaijeveld, Guay, Mo, Lawson, McGrath, Vendrow, Jiao, Lee, Steele, Wang,
  Mao, Chen, Hayashi, Xiao, Salahi, Wu, Sekhri, Sharma, Singhal, Li, Nguyen,
  Gu-Lemberg, King, Liu, Stone, Yu, Ying, Georgiev, Lim, Tirumala, Miller,
  Ahmad, Lv, Clare, Fauconnet, Itow, Yang, Romaniuk, Anise, Byron, Pathak,
  Maksin, Lo, Ho, Jing, Wu, Xiong, Mamitsuka, Yang, McCallum, Held, Bourgeois,
  Engstrom, Kuhn, Feuvrier, Zhang, Switzer, Kondraciuk, Kaiser, Joglekar,
  Singh, Shah, Stratta, Williams, Chen, Sun, Cayton, Li, Zhang, Aljubeh,
  Nichols, Haines, Schwarzer, Gupta, Shah, Guan, Huang, Dong, Wang, Glaese,
  Carroll, Lampe, Malek, Sharman, Zhang, Wang, Pokrass, Florian, Pavlov, Wang,
  Chen, Wang, Feng, Bavarian, Lin, Abdool, Rohaninejad, Soto, Staudacher,
  LaFontaine, Marwell, Liu, Preston, Turley, Ansman, Blades, Pancha, Mikhaylin,
  Felix, Handa, Rai, Keskar, Brown, Nachum, Boiko, Murk, Watkins, Gleeson,
  Mishkin, Lesiewicz, Baltescu, Belov, Zhokhov, Pronin, Guo, Thacker, Liu,
  Yuan, Liu, Dias, Puckett, Arora, Mullapudi, Gaon, Miyara, Song, Aggarwal,
  Marsan, Yemiru, Xiong, Kshirsagar, Nuttall, Tsiupa, Eldan, Wang, James, Ziv,
  Shu, Nigmatullin, Jain, Talaie, Altman, Arnesen, Toizer, Toyer, Miserendino,
  Agarwal, Yoo, Heon, Ethersmith, Grove, Taylor, Bubeck, Banesiu, Amdo, Zhao,
  Wu, Santurkar, Zhao, Chaudhuri, Krishnaswamy, Shuaiqi, Xia, Cheng, Anadkat,
  Fishman, Tobin, Fu, Jain, Mei, Egoian, Kim, Golden, Mah, Lin, Imm, Sharpe,
  Yadlowsky, Choudhry, Eum, Sanjeev, Khan, Stramer, Wang, Xin, Gogineni,
  Christianson, Sanders, Patwardhan, Degry, Shadwell, Fu, Gao, Garipov,
  Sriskandarajah, Sherbakov, Korbak, Kaftan, Hiratsuka, Wang, Song, Zhao,
  Peterson, Kharitonov, Chernova, Kosaraju, Kuo, Pong, Verma, Petrov, Jiang,
  Zhang, Zhou, Xie, Zhan, McCabe, DePue, Ellsworth, Bain, Thompson, Chen, Qi,
  Xiang, Shi, Dubois, Yu, Khakbaz, Wu, Qian, Lee, Chen, Zhang, Xiong, Tian,
  Cha, Bai, Yang, Yuan, Li, Zhang, Yang, Jin, Jiang, Wang, Wang, Liu,
  Stubenvoll, Dou, Wu, and Wang]{singh2026openaigpt5card}
Aaditya Singh, Adam Fry, Adam Perelman, Adam Tart, Adi Ganesh, Ahmed El-Kishky,
  Aidan McLaughlin, Aiden Low, AJ~Ostrow, Akhila Ananthram, Akshay Nathan, Alan
  Luo, Alec Helyar, Aleksander Madry, Aleksandr Efremov, Aleksandra Spyra, Alex
  Baker-Whitcomb, Alex Beutel, Alex Karpenko, Alex Makelov, Alex Neitz, Alex
  Wei, Alexandra Barr, Alexandre Kirchmeyer, Alexey Ivanov, Alexi Christakis,
  Alistair Gillespie, Allison Tam, Ally Bennett, Alvin Wan, Alyssa Huang,
  Amy~McDonald Sandjideh, Amy Yang, Ananya Kumar, Andre Saraiva, Andrea
  Vallone, Andrei Gheorghe, Andres~Garcia Garcia, Andrew Braunstein, Andrew
  Liu, Andrew Schmidt, Andrey Mereskin, Andrey Mishchenko, Andy Applebaum, Andy
  Rogerson, Ann Rajan, Annie Wei, Anoop Kotha, Anubha Srivastava, Anushree
  Agrawal, Arun Vijayvergiya, Ashley Tyra, Ashvin Nair, Avi Nayak, Ben Eggers,
  Bessie Ji, Beth Hoover, Bill Chen, Blair Chen, Boaz Barak, Borys Minaiev,
  Botao Hao, Bowen Baker, Brad Lightcap, Brandon McKinzie, Brandon Wang,
  Brendan Quinn, Brian Fioca, Brian Hsu, Brian Yang, Brian Yu, Brian Zhang,
  Brittany Brenner, Callie~Riggins Zetino, Cameron Raymond, Camillo Lugaresi,
  Carolina Paz, Cary Hudson, Cedric Whitney, Chak Li, Charles Chen, Charlotte
  Cole, Chelsea Voss, Chen Ding, Chen Shen, Chengdu Huang, Chris Colby, Chris
  Hallacy, Chris Koch, Chris Lu, Christina Kaplan, Christina Kim,
  CJ~Minott-Henriques, Cliff Frey, Cody Yu, Coley Czarnecki, Colin Reid, Colin
  Wei, Cory Decareaux, Cristina Scheau, Cyril Zhang, Cyrus Forbes, Da~Tang,
  Dakota Goldberg, Dan Roberts, Dana Palmie, Daniel Kappler, Daniel Levine,
  Daniel Wright, Dave Leo, David Lin, David Robinson, Declan Grabb, Derek Chen,
  Derek Lim, Derek Salama, Dibya Bhattacharjee, Dimitris Tsipras, Dinghua Li,
  Dingli Yu, DJ~Strouse, Drew Williams, Dylan Hunn, Ed~Bayes, Edwin Arbus, Ekin
  Akyurek, Elaine~Ya Le, Elana Widmann, Eli Yani, Elizabeth Proehl, Enis Sert,
  Enoch Cheung, Eri Schwartz, Eric Han, Eric Jiang, Eric Mitchell, Eric Sigler,
  Eric Wallace, Erik Ritter, Erin Kavanaugh, Evan Mays, Evgenii Nikishin,
  Fangyuan Li, Felipe~Petroski Such, Filipe de~Avila Belbute~Peres, Filippo
  Raso, Florent Bekerman, Foivos Tsimpourlas, Fotis Chantzis, Francis Song,
  Francis Zhang, Gaby Raila, Garrett McGrath, Gary Briggs, Gary Yang,
  Giambattista Parascandolo, Gildas Chabot, Grace Kim, Grace Zhao, Gregory
  Valiant, Guillaume Leclerc, Hadi Salman, Hanson Wang, Hao Sheng, Haoming
  Jiang, Haoyu Wang, Haozhun Jin, Harshit Sikchi, Heather Schmidt, Henry
  Aspegren, Honglin Chen, Huida Qiu, Hunter Lightman, Ian Covert, Ian
  Kivlichan, Ian Silber, Ian Sohl, Ibrahim Hammoud, Ignasi Clavera, Ikai Lan,
  Ilge Akkaya, Ilya Kostrikov, Irina Kofman, Isak Etinger, Ishaan Singal,
  Jackie Hehir, Jacob Huh, Jacqueline Pan, Jake Wilczynski, Jakub Pachocki,
  James Lee, James Quinn, Jamie Kiros, Janvi Kalra, Jasmyn Samaroo, Jason Wang,
  Jason Wolfe, Jay Chen, Jay Wang, Jean Harb, Jeffrey Han, Jeffrey Wang,
  Jennifer Zhao, Jeremy Chen, Jerene Yang, Jerry Tworek, Jesse Chand, Jessica
  Landon, Jessica Liang, Ji~Lin, Jiancheng Liu, Jianfeng Wang, Jie Tang, Jihan
  Yin, Joanne Jang, Joel Morris, Joey Flynn, Johannes Ferstad, Johannes
  Heidecke, John Fishbein, John Hallman, Jonah Grant, Jonathan Chien, Jonathan
  Gordon, Jongsoo Park, Jordan Liss, Jos Kraaijeveld, Joseph Guay, Joseph Mo,
  Josh Lawson, Josh McGrath, Joshua Vendrow, Joy Jiao, Julian Lee, Julie
  Steele, Julie Wang, Junhua Mao, Kai Chen, Kai Hayashi, Kai Xiao, Kamyar
  Salahi, Kan Wu, Karan Sekhri, Karan Sharma, Karan Singhal, Karen Li, Kenny
  Nguyen, Keren Gu-Lemberg, Kevin King, Kevin Liu, Kevin Stone, Kevin Yu,
  Kristen Ying, Kristian Georgiev, Kristie Lim, Kushal Tirumala, Kyle Miller,
  Lama Ahmad, Larry Lv, Laura Clare, Laurance Fauconnet, Lauren Itow, Lauren
  Yang, Laurentia Romaniuk, Leah Anise, Lee Byron, Leher Pathak, Leon Maksin,
  Leyan Lo, Leyton Ho, Li~Jing, Liang Wu, Liang Xiong, Lien Mamitsuka, Lin
  Yang, Lindsay McCallum, Lindsey Held, Liz Bourgeois, Logan Engstrom, Lorenz
  Kuhn, Louis Feuvrier, Lu~Zhang, Lucas Switzer, Lukas Kondraciuk, Lukasz
  Kaiser, Manas Joglekar, Mandeep Singh, Mandip Shah, Manuka Stratta, Marcus
  Williams, Mark Chen, Mark Sun, Marselus Cayton, Martin Li, Marvin Zhang,
  Marwan Aljubeh, Matt Nichols, Matthew Haines, Max Schwarzer, Mayank Gupta,
  Meghan Shah, Melody~Y. Guan, Melody Huang, Meng Dong, Mengqing Wang, Mia
  Glaese, Micah Carroll, Michael Lampe, Michael Malek, Michael Sharman, Michael
  Zhang, Michele Wang, Michelle Pokrass, Mihai Florian, Mikhail Pavlov, Miles
  Wang, Ming Chen, Mingxuan Wang, Minnia Feng, Mo~Bavarian, Molly Lin, Moose
  Abdool, Mostafa Rohaninejad, Nacho Soto, Natalie Staudacher, Natan
  LaFontaine, Nathan Marwell, Nelson Liu, Nick Preston, Nick Turley, Nicklas
  Ansman, Nicole Blades, Nikil Pancha, Nikita Mikhaylin, Niko Felix, Nikunj
  Handa, Nishant Rai, Nitish Keskar, Noam Brown, Ofir Nachum, Oleg Boiko, Oleg
  Murk, Olivia Watkins, Oona Gleeson, Pamela Mishkin, Patryk Lesiewicz, Paul
  Baltescu, Pavel Belov, Peter Zhokhov, Philip Pronin, Phillip Guo, Phoebe
  Thacker, Qi~Liu, Qiming Yuan, Qinghua Liu, Rachel Dias, Rachel Puckett, Rahul
  Arora, Ravi~Teja Mullapudi, Raz Gaon, Reah Miyara, Rennie Song, Rishabh
  Aggarwal, RJ~Marsan, Robel Yemiru, Robert Xiong, Rohan Kshirsagar, Rohan
  Nuttall, Roman Tsiupa, Ronen Eldan, Rose Wang, Roshan James, Roy Ziv, Rui
  Shu, Ruslan Nigmatullin, Saachi Jain, Saam Talaie, Sam Altman, Sam Arnesen,
  Sam Toizer, Sam Toyer, Samuel Miserendino, Sandhini Agarwal, Sarah Yoo,
  Savannah Heon, Scott Ethersmith, Sean Grove, Sean Taylor, Sebastien Bubeck,
  Sever Banesiu, Shaokyi Amdo, Shengjia Zhao, Sherwin Wu, Shibani Santurkar,
  Shiyu Zhao, Shraman~Ray Chaudhuri, Shreyas Krishnaswamy, Shuaiqi, Xia,
  Shuyang Cheng, Shyamal Anadkat, Simón~Posada Fishman, Simon Tobin, Siyuan
  Fu, Somay Jain, Song Mei, Sonya Egoian, Spencer Kim, Spug Golden, SQ~Mah,
  Steph Lin, Stephen Imm, Steve Sharpe, Steve Yadlowsky, Sulman Choudhry,
  Sungwon Eum, Suvansh Sanjeev, Tabarak Khan, Tal Stramer, Tao Wang, Tao Xin,
  Tarun Gogineni, Taya Christianson, Ted Sanders, Tejal Patwardhan, Thomas
  Degry, Thomas Shadwell, Tianfu Fu, Tianshi Gao, Timur Garipov, Tina
  Sriskandarajah, Toki Sherbakov, Tomek Korbak, Tomer Kaftan, Tomo Hiratsuka,
  Tongzhou Wang, Tony Song, Tony Zhao, Troy Peterson, Val Kharitonov, Victoria
  Chernova, Vineet Kosaraju, Vishal Kuo, Vitchyr Pong, Vivek Verma, Vlad
  Petrov, Wanning Jiang, Weixing Zhang, Wenda Zhou, Wenlei Xie, Wenting Zhan,
  Wes McCabe, Will DePue, Will Ellsworth, Wulfie Bain, Wyatt Thompson,
  Xiangning Chen, Xiangyu Qi, Xin Xiang, Xinwei Shi, Yann Dubois, Yaodong Yu,
  Yara Khakbaz, Yifan Wu, Yilei Qian, Yin~Tat Lee, Yinbo Chen, Yizhen Zhang,
  Yizhong Xiong, Yonglong Tian, Young Cha, Yu~Bai, Yu~Yang, Yuan Yuan, Yuanzhi
  Li, Yufeng Zhang, Yuguang Yang, Yujia Jin, Yun Jiang, Yunyun Wang, Yushi
  Wang, Yutian Liu, Zach Stubenvoll, Zehao Dou, Zheng Wu, and Zhigang Wang.
\newblock Openai gpt-5 system card, 2026.
\newblock URL \url{https://arxiv.org/abs/2601.03267}.

\bibitem[Tarek Ibn~Ziad and Kozyrakis(2026)]{tarek2026hunting}
Mohamed Tarek Ibn~Ziad and Christos Kozyrakis.
\newblock Hunting cuda bugs at scale with cufuzz.
\newblock \emph{Proceedings of the ACM on Programming Languages}, 10\penalty0
  (OOPSLA1):\penalty0 877--904, 2026.

\bibitem[Tarek Ibn~Ziad et~al.(2023)Tarek Ibn~Ziad, Damani, Jaleel, Keckler,
  and Stephenson]{tarek2023cucatch}
Mohamed Tarek Ibn~Ziad, Sana Damani, Aamer Jaleel, Stephen~W Keckler, and Mark
  Stephenson.
\newblock Cucatch: A debugging tool for efficiently catching memory safety
  violations in cuda applications.
\newblock \emph{Proceedings of the ACM on Programming Languages}, 7\penalty0
  (PLDI):\penalty0 124--147, 2023.

\bibitem[Team et~al.(2025{\natexlab{a}})Team, Zeng, Lv, Zheng, Hou, Chen, Xie,
  Wang, Yin, Zeng, Zhang, Wang, Zhong, Liu, Lu, Cao, Zhang, Huang, Wei, Cheng,
  An, Niu, Wen, Bai, Du, Wang, Zhu, Zhang, Wen, Wu, Xu, Huang, Zhao, Cai, Yu,
  Li, Ge, Huang, Zhang, Xu, Zhu, Li, Yin, Lin, Yang, Jiang, Ai, Zhu, Wang, Pan,
  Wang, Sun, Li, Li, Hu, Zhang, Peng, Tai, Zhang, Wang, Yang, Liu, Zhao, Liu,
  Yan, Liu, Chen, Li, Zhao, Ren, Jiao, Zhao, Yan, Wang, Gui, Zhao, Liu, Li, Li,
  Lu, Wang, Yuan, Li, Du, Du, Liu, Zhi, Gao, Wang, Yang, Xu, Fan, Wu, Ding,
  Wang, Zhang, Li, Xu, Zhao, Zhai, Du, Dong, Lei, Tu, Yang, Lu, Li, Li,
  Shuang-Li, Yang, Yi, Yu, Tian, Wang, Yu, Tam, Liang, Liu, Wang, Jia, Gu,
  Ling, Wang, Fan, Pan, Zhang, Zhang, Fu, Zhang, Xu, Wu, Lu, Wang, Zhou, Pan,
  Zhang, Wang, Li, Su, Geng, Zhu, Yang, Li, Wu, Li, Liu, Wang, Li, Zhang, Liu,
  Yang, Zhou, Qiao, Feng, Liu, Zhang, Wang, Yao, Wang, Liu, Chai, Li, Zhao,
  Chen, Zhai, Xu, Huang, Wang, Li, Dong, and
  Tang]{5team2025glm45agenticreasoningcoding}
5~Team, Aohan Zeng, Xin Lv, Qinkai Zheng, Zhenyu Hou, Bin Chen, Chengxing Xie,
  Cunxiang Wang, Da~Yin, Hao Zeng, Jiajie Zhang, Kedong Wang, Lucen Zhong,
  Mingdao Liu, Rui Lu, Shulin Cao, Xiaohan Zhang, Xuancheng Huang, Yao Wei,
  Yean Cheng, Yifan An, Yilin Niu, Yuanhao Wen, Yushi Bai, Zhengxiao Du, Zihan
  Wang, Zilin Zhu, Bohan Zhang, Bosi Wen, Bowen Wu, Bowen Xu, Can Huang, Casey
  Zhao, Changpeng Cai, Chao Yu, Chen Li, Chendi Ge, Chenghua Huang, Chenhui
  Zhang, Chenxi Xu, Chenzheng Zhu, Chuang Li, Congfeng Yin, Daoyan Lin, Dayong
  Yang, Dazhi Jiang, Ding Ai, Erle Zhu, Fei Wang, Gengzheng Pan, Guo Wang,
  Hailong Sun, Haitao Li, Haiyang Li, Haiyi Hu, Hanyu Zhang, Hao Peng, Hao Tai,
  Haoke Zhang, Haoran Wang, Haoyu Yang, He~Liu, He~Zhao, Hongwei Liu, Hongxi
  Yan, Huan Liu, Huilong Chen, Ji~Li, Jiajing Zhao, Jiamin Ren, Jian Jiao,
  Jiani Zhao, Jianyang Yan, Jiaqi Wang, Jiayi Gui, Jiayue Zhao, Jie Liu, Jijie
  Li, Jing Li, Jing Lu, Jingsen Wang, Jingwei Yuan, Jingxuan Li, Jingzhao Du,
  Jinhua Du, Jinxin Liu, Junkai Zhi, Junli Gao, Ke~Wang, Lekang Yang, Liang Xu,
  Lin Fan, Lindong Wu, Lintao Ding, Lu~Wang, Man Zhang, Minghao Li, Minghuan
  Xu, Mingming Zhao, Mingshu Zhai, Pengfan Du, Qian Dong, Shangde Lei,
  Shangqing Tu, Shangtong Yang, Shaoyou Lu, Shijie Li, Shuang Li, Shuang-Li,
  Shuxun Yang, Sibo Yi, Tianshu Yu, Wei Tian, Weihan Wang, Wenbo Yu, Weng~Lam
  Tam, Wenjie Liang, Wentao Liu, Xiao Wang, Xiaohan Jia, Xiaotao Gu, Xiaoying
  Ling, Xin Wang, Xing Fan, Xingru Pan, Xinyuan Zhang, Xinze Zhang, Xiuqing Fu,
  Xunkai Zhang, Yabo Xu, Yandong Wu, Yida Lu, Yidong Wang, Yilin Zhou, Yiming
  Pan, Ying Zhang, Yingli Wang, Yingru Li, Yinpei Su, Yipeng Geng, Yitong Zhu,
  Yongkun Yang, Yuhang Li, Yuhao Wu, Yujiang Li, Yunan Liu, Yunqing Wang,
  Yuntao Li, Yuxuan Zhang, Zezhen Liu, Zhen Yang, Zhengda Zhou, Zhongpei Qiao,
  Zhuoer Feng, Zhuorui Liu, Zichen Zhang, Zihan Wang, Zijun Yao, Zikang Wang,
  Ziqiang Liu, Ziwei Chai, Zixuan Li, Zuodong Zhao, Wenguang Chen, Jidong Zhai,
  Bin Xu, Minlie Huang, Hongning Wang, Juanzi Li, Yuxiao Dong, and Jie Tang.
\newblock Glm-4.5: Agentic, reasoning, and coding (arc) foundation models,
  2025{\natexlab{a}}.
\newblock URL \url{https://arxiv.org/abs/2508.06471}.

\bibitem[Team et~al.(2025{\natexlab{b}})Team, Kamath, Ferret, Pathak,
  Vieillard, Merhej, Perrin, Matejovicova, Ramé, Rivière, Rouillard, Mesnard,
  Cideron, bastien Grill, Ramos, Yvinec, Casbon, Pot, Penchev, Liu, Visin,
  Kenealy, Beyer, Zhai, Tsitsulin, Busa-Fekete, Feng, Sachdeva, Coleman, Gao,
  Mustafa, Barr, Parisotto, Tian, Eyal, Cherry, Peter, Sinopalnikov,
  Bhupatiraju, Agarwal, Kazemi, Malkin, Kumar, Vilar, Brusilovsky, Luo,
  Steiner, Friesen, Sharma, Sharma, Gilady, Goedeckemeyer, Saade, Feng,
  Kolesnikov, Bendebury, Abdagic, Vadi, György, Pinto, Das, Bapna, Miech,
  Yang, Paterson, Shenoy, Chakrabarti, Piot, Wu, Shahriari, Petrini, Chen, Lan,
  Choquette-Choo, Carey, Brick, Deutsch, Eisenbud, Cattle, Cheng, Paparas,
  Sreepathihalli, Reid, Tran, Zelle, Noland, Huizenga, Kharitonov, Liu,
  Amirkhanyan, Cameron, Hashemi, Klimczak-Plucińska, Singh, Mehta, Lehri,
  Hazimeh, Ballantyne, Szpektor, Nardini, Pouget-Abadie, Chan, Stanton,
  Wieting, Lai, Orbay, Fernandez, Newlan, yeong Ji, Singh, Black, Yu, Hui,
  Vodrahalli, Greff, Qiu, Valentine, Coelho, Ritter, Hoffman, Watson,
  Chaturvedi, Moynihan, Ma, Babar, Noy, Byrd, Roy, Momchev, Chauhan, Sachdeva,
  Bunyan, Botarda, Caron, Rubenstein, Culliton, Schmid, Sessa, Xu, Stanczyk,
  Tafti, Shivanna, Wu, Pan, Rokni, Willoughby, Vallu, Mullins, Jerome, Smoot,
  Girgin, Iqbal, Reddy, Sheth, Põder, Bhatnagar, Panyam, Eiger, Zhang, Liu,
  Yacovone, Liechty, Kalra, Evci, Misra, Roseberry, Feinberg, Kolesnikov, Han,
  Kwon, Chen, Chow, Zhu, Wei, Egyed, Cotruta, Giang, Kirk, Rao, Black, Babar,
  Lo, Moreira, Martins, Sanseviero, Gonzalez, Gleicher, Warkentin, Mirrokni,
  Senter, Collins, Barral, Ghahramani, Hadsell, Matias, Sculley, Petrov,
  Fiedel, Shazeer, Vinyals, Dean, Hassabis, Kavukcuoglu, Farabet, Buchatskaya,
  Alayrac, Anil, Dmitry, Lepikhin, Borgeaud, Bachem, Joulin, Andreev, Hardin,
  Dadashi, and Hussenot]{gemmateam2025gemma3technicalreport}
Gemma Team, Aishwarya Kamath, Johan Ferret, Shreya Pathak, Nino Vieillard,
  Ramona Merhej, Sarah Perrin, Tatiana Matejovicova, Alexandre Ramé, Morgane
  Rivière, Louis Rouillard, Thomas Mesnard, Geoffrey Cideron, Jean bastien
  Grill, Sabela Ramos, Edouard Yvinec, Michelle Casbon, Etienne Pot, Ivo
  Penchev, Gaël Liu, Francesco Visin, Kathleen Kenealy, Lucas Beyer, Xiaohai
  Zhai, Anton Tsitsulin, Robert Busa-Fekete, Alex Feng, Noveen Sachdeva,
  Benjamin Coleman, Yi~Gao, Basil Mustafa, Iain Barr, Emilio Parisotto, David
  Tian, Matan Eyal, Colin Cherry, Jan-Thorsten Peter, Danila Sinopalnikov,
  Surya Bhupatiraju, Rishabh Agarwal, Mehran Kazemi, Dan Malkin, Ravin Kumar,
  David Vilar, Idan Brusilovsky, Jiaming Luo, Andreas Steiner, Abe Friesen,
  Abhanshu Sharma, Abheesht Sharma, Adi~Mayrav Gilady, Adrian Goedeckemeyer,
  Alaa Saade, Alex Feng, Alexander Kolesnikov, Alexei Bendebury, Alvin Abdagic,
  Amit Vadi, András György, André~Susano Pinto, Anil Das, Ankur Bapna,
  Antoine Miech, Antoine Yang, Antonia Paterson, Ashish Shenoy, Ayan
  Chakrabarti, Bilal Piot, Bo~Wu, Bobak Shahriari, Bryce Petrini, Charlie Chen,
  Charline~Le Lan, Christopher~A. Choquette-Choo, CJ~Carey, Cormac Brick,
  Daniel Deutsch, Danielle Eisenbud, Dee Cattle, Derek Cheng, Dimitris Paparas,
  Divyashree~Shivakumar Sreepathihalli, Doug Reid, Dustin Tran, Dustin Zelle,
  Eric Noland, Erwin Huizenga, Eugene Kharitonov, Frederick Liu, Gagik
  Amirkhanyan, Glenn Cameron, Hadi Hashemi, Hanna Klimczak-Plucińska, Harman
  Singh, Harsh Mehta, Harshal~Tushar Lehri, Hussein Hazimeh, Ian Ballantyne,
  Idan Szpektor, Ivan Nardini, Jean Pouget-Abadie, Jetha Chan, Joe Stanton,
  John Wieting, Jonathan Lai, Jordi Orbay, Joseph Fernandez, Josh Newlan,
  Ju~yeong Ji, Jyotinder Singh, Kat Black, Kathy Yu, Kevin Hui, Kiran
  Vodrahalli, Klaus Greff, Linhai Qiu, Marcella Valentine, Marina Coelho,
  Marvin Ritter, Matt Hoffman, Matthew Watson, Mayank Chaturvedi, Michael
  Moynihan, Min Ma, Nabila Babar, Natasha Noy, Nathan Byrd, Nick Roy, Nikola
  Momchev, Nilay Chauhan, Noveen Sachdeva, Oskar Bunyan, Pankil Botarda, Paul
  Caron, Paul~Kishan Rubenstein, Phil Culliton, Philipp Schmid, Pier~Giuseppe
  Sessa, Pingmei Xu, Piotr Stanczyk, Pouya Tafti, Rakesh Shivanna, Renjie Wu,
  Renke Pan, Reza Rokni, Rob Willoughby, Rohith Vallu, Ryan Mullins, Sammy
  Jerome, Sara Smoot, Sertan Girgin, Shariq Iqbal, Shashir Reddy, Shruti Sheth,
  Siim Põder, Sijal Bhatnagar, Sindhu~Raghuram Panyam, Sivan Eiger, Susan
  Zhang, Tianqi Liu, Trevor Yacovone, Tyler Liechty, Uday Kalra, Utku Evci,
  Vedant Misra, Vincent Roseberry, Vlad Feinberg, Vlad Kolesnikov, Woohyun Han,
  Woosuk Kwon, Xi~Chen, Yinlam Chow, Yuvein Zhu, Zichuan Wei, Zoltan Egyed,
  Victor Cotruta, Minh Giang, Phoebe Kirk, Anand Rao, Kat Black, Nabila Babar,
  Jessica Lo, Erica Moreira, Luiz~Gustavo Martins, Omar Sanseviero, Lucas
  Gonzalez, Zach Gleicher, Tris Warkentin, Vahab Mirrokni, Evan Senter, Eli
  Collins, Joelle Barral, Zoubin Ghahramani, Raia Hadsell, Yossi Matias,
  D.~Sculley, Slav Petrov, Noah Fiedel, Noam Shazeer, Oriol Vinyals, Jeff Dean,
  Demis Hassabis, Koray Kavukcuoglu, Clement Farabet, Elena Buchatskaya,
  Jean-Baptiste Alayrac, Rohan Anil, Dmitry, Lepikhin, Sebastian Borgeaud,
  Olivier Bachem, Armand Joulin, Alek Andreev, Cassidy Hardin, Robert Dadashi,
  and Léonard Hussenot.
\newblock Gemma 3 technical report, 2025{\natexlab{b}}.
\newblock URL \url{https://arxiv.org/abs/2503.19786}.

\bibitem[Team et~al.(2026)Team, Bai, Bao, Charles, Chen, Chen, Chen, Chen,
  Chen, Chen, Chen, Chen, Chen, Chen, Chen, Cui, Ding, Dong, Du, Du, Du, Du,
  Fan, Feng, Fu, Gao, Gao, Gao, Gao, Gao, Ge, Geng, Gu, Gu, Guan, Guo, Guo,
  Hao, He, He, He, He, Hong, Hu, Hu, Hu, Huang, Huang, Huang, Jiang, Jiang,
  Jin, Kang, Lai, Li, Li, Li, Li, Li, Li, Li, Li, Li, Li, Lin, Lin, Lin, Liu,
  Liu, Liu, Liu, Liu, Liu, Liu, Liu, Liu, Liu, Liu, Liu, Liu, Liu, Liu, Lu, Lu,
  Lu, Luo, Ma, Ma, Ma, Mao, Mei, Men, Miao, Pan, Peng, Qin, Qin, Qu, Shang,
  Shi, Shi, Song, Su, Su, Sui, Sun, Sung, Tai, Tang, Tao, Teng, Tian, Wang,
  Wang, Wang, Wang, Wang, Wang, Wang, Wang, Wang, Wang, Wang, Wang, Wang, Wang,
  Wang, Wang, Wang, Wang, Wang, Wang, Wang, Wei, Wei, Wu, Wu, Wu, Wu, Xiao,
  Xie, Xie, Xiong, Xu, Xu, Xu, Xu, Xu, Xu, Xu, Xu, Xu, Xu, Xu, Yan, Yan, Yang,
  Yang, Yang, Yang, Yang, Yang, Yang, Yao, Yao, Ye, Ye, Yin, Yu, Yuan, Yuan,
  Yuan, Yuan, Zhan, Zhang, Zhang, Zhang, Zhang, Zhang, Zhang, Zhang, Zhang,
  Zhang, Zhang, Zhang, Zhang, Zhang, Zhao, Zhao, Zhao, Zheng, Zheng, Zhong,
  Zhou, Zhou, Zhou, Zhu, Zhu, Zhuang, and Zu]{kimiteam2026kimik2openagentic}
Kimi Team, Yifan Bai, Yiping Bao, Y.~Charles, Cheng Chen, Guanduo Chen, Haiting
  Chen, Huarong Chen, Jiahao Chen, Ningxin Chen, Ruijue Chen, Yanru Chen,
  Yuankun Chen, Yutian Chen, Zhuofu Chen, Jialei Cui, Hao Ding, Mengnan Dong,
  Angang Du, Chenzhuang Du, Dikang Du, Yulun Du, Yu~Fan, Yichen Feng, Kelin Fu,
  Bofei Gao, Chenxiao Gao, Hongcheng Gao, Peizhong Gao, Tong Gao, Yuyao Ge,
  Shangyi Geng, Qizheng Gu, Xinran Gu, Longyu Guan, Haiqing Guo, Jianhang Guo,
  Xiaoru Hao, Tianhong He, Weiran He, Wenyang He, Yunjia He, Chao Hong, Hao Hu,
  Yangyang Hu, Zhenxing Hu, Weixiao Huang, Zhiqi Huang, Zihao Huang, Tao Jiang,
  Zhejun Jiang, Xinyi Jin, Yongsheng Kang, Guokun Lai, Cheng Li, Fang Li,
  Haoyang Li, Ming Li, Wentao Li, Yang Li, Yanhao Li, Yiwei Li, Zhaowei Li,
  Zheming Li, Hongzhan Lin, Xiaohan Lin, Zongyu Lin, Chengyin Liu, Chenyu Liu,
  Hongzhang Liu, Jingyuan Liu, Junqi Liu, Liang Liu, Shaowei Liu, T.~Y. Liu,
  Tianwei Liu, Weizhou Liu, Yangyang Liu, Yibo Liu, Yiping Liu, Yue Liu,
  Zhengying Liu, Enzhe Lu, Haoyu Lu, Lijun Lu, Yashuo Luo, Shengling Ma, Xinyu
  Ma, Yingwei Ma, Shaoguang Mao, Jie Mei, Xin Men, Yibo Miao, Siyuan Pan, Yebo
  Peng, Ruoyu Qin, Zeyu Qin, Bowen Qu, Zeyu Shang, Lidong Shi, Shengyuan Shi,
  Feifan Song, Jianlin Su, Zhengyuan Su, Lin Sui, Xinjie Sun, Flood Sung,
  Yunpeng Tai, Heyi Tang, Jiawen Tao, Qifeng Teng, Chaoran Tian, Chensi Wang,
  Dinglu Wang, Feng Wang, Hailong Wang, Haiming Wang, Jianzhou Wang, Jiaxing
  Wang, Jinhong Wang, Shengjie Wang, Shuyi Wang, Si~Wang, Xinyuan Wang, Yao
  Wang, Yejie Wang, Yiqin Wang, Yuxin Wang, Yuzhi Wang, Zhaoji Wang, Zhengtao
  Wang, Zhengtao Wang, Zhexu Wang, Chu Wei, Qianqian Wei, Haoning Wu, Wenhao
  Wu, Xingzhe Wu, Yuxin Wu, Chenjun Xiao, Jin Xie, Xiaotong Xie, Weimin Xiong,
  Boyu Xu, Jinjing Xu, L.~H. Xu, Lin Xu, Suting Xu, Weixin Xu, Xinran Xu,
  Yangchuan Xu, Ziyao Xu, Jing Xu, Jing Xu, Junjie Yan, Yuzi Yan, Hao Yang,
  Xiaofei Yang, Yi~Yang, Ying Yang, Zhen Yang, Zhilin Yang, Zonghan Yang,
  Haotian Yao, Xingcheng Yao, Wenjie Ye, Zhuorui Ye, Bohong Yin, Longhui Yu,
  Enming Yuan, Hongbang Yuan, Mengjie Yuan, Siyu Yuan, Haobing Zhan, Dehao
  Zhang, Hao Zhang, Wanlu Zhang, Xiaobin Zhang, Yadong Zhang, Yangkun Zhang,
  Yichi Zhang, Yizhi Zhang, Yongting Zhang, Yu~Zhang, Yutao Zhang, Yutong
  Zhang, Zheng Zhang, Haotian Zhao, Yikai Zhao, Zijia Zhao, Huabin Zheng,
  Shaojie Zheng, Longguang Zhong, Jianren Zhou, Xinyu Zhou, Zaida Zhou, Jinguo
  Zhu, Zhen Zhu, Weiyu Zhuang, and Xinxing Zu.
\newblock Kimi k2: Open agentic intelligence, 2026.
\newblock URL \url{https://arxiv.org/abs/2507.20534}.

\bibitem[{ThunderKittens Authors}(2026)]{thunderkittensrepo}
{ThunderKittens Authors}.
\newblock Thunderkittens.
\newblock GitHub repository, 2026.
\newblock URL \url{https://github.com/HazyResearch/ThunderKittens}.

\bibitem[Tian et~al.(2024)Tian, Ye, Qin, Cong, Lin, Pan, Wu, Haotian, Weichuan,
  Liu, et~al.]{tian2024debugbench}
Runchu Tian, Yining Ye, Yujia Qin, Xin Cong, Yankai Lin, Yinxu Pan, Yesai Wu,
  Hui Haotian, Liu Weichuan, Zhiyuan Liu, et~al.
\newblock Debugbench: Evaluating debugging capability of large language models.
\newblock In \emph{Findings of the Association for Computational Linguistics:
  ACL 2024}, pages 4173--4198, 2024.

\bibitem[Wu et~al.(2019)Wu, Zhou, Zhang, Liu, and Zhang]{wu2019characterizing}
Mingyuan Wu, Husheng Zhou, Lingming Zhang, Cong Liu, and Yuqun Zhang.
\newblock Characterizing and detecting cuda program bugs.
\newblock \emph{arXiv preprint arXiv:1905.01833}, 2019.

\bibitem[Xia and Zhang(2023)]{xia2023keep}
Chunqiu~Steven Xia and Lingming Zhang.
\newblock Keep the conversation going: Fixing 162 out of 337 bugs for \$0.42
  each using chatgpt.
\newblock \emph{arXiv preprint arXiv:2304.00385}, 2023.

\bibitem[Yang et~al.(2025)Yang, Li, Yang, Zhang, Hui, Zheng, Yu, Gao, Huang,
  Lv, Zheng, Liu, Zhou, Huang, Hu, Ge, Wei, Lin, Tang, Yang, Tu, Zhang, Yang,
  Yang, Zhou, Zhou, Lin, Dang, Bao, Yang, Yu, Deng, Li, Xue, Li, Zhang, Wang,
  Zhu, Men, Gao, Liu, Luo, Li, Tang, Yin, Ren, Wang, Zhang, Ren, Fan, Su,
  Zhang, Zhang, Wan, Liu, Wang, Cui, Zhang, Zhou, and
  Qiu]{yang2025qwen3technicalreport}
An~Yang, Anfeng Li, Baosong Yang, Beichen Zhang, Binyuan Hui, Bo~Zheng, Bowen
  Yu, Chang Gao, Chengen Huang, Chenxu Lv, Chujie Zheng, Dayiheng Liu, Fan
  Zhou, Fei Huang, Feng Hu, Hao Ge, Haoran Wei, Huan Lin, Jialong Tang, Jian
  Yang, Jianhong Tu, Jianwei Zhang, Jianxin Yang, Jiaxi Yang, Jing Zhou,
  Jingren Zhou, Junyang Lin, Kai Dang, Keqin Bao, Kexin Yang, Le~Yu, Lianghao
  Deng, Mei Li, Mingfeng Xue, Mingze Li, Pei Zhang, Peng Wang, Qin Zhu, Rui
  Men, Ruize Gao, Shixuan Liu, Shuang Luo, Tianhao Li, Tianyi Tang, Wenbiao
  Yin, Xingzhang Ren, Xinyu Wang, Xinyu Zhang, Xuancheng Ren, Yang Fan, Yang
  Su, Yichang Zhang, Yinger Zhang, Yu~Wan, Yuqiong Liu, Zekun Wang, Zeyu Cui,
  Zhenru Zhang, Zhipeng Zhou, and Zihan Qiu.
\newblock Qwen3 technical report, 2025.
\newblock URL \url{https://arxiv.org/abs/2505.09388}.

\bibitem[Zhang et~al.(2025)Zhang, Wang, Li, Luo, Hong, and Ding]{cudaforge}
Zijian Zhang, Rong Wang, Shiyang Li, Yuebo Luo, Mingyi Hong, and Caiwen Ding.
\newblock Cudaforge: An agent framework with hardware feedback for cuda kernel
  optimization, 2025.

\bibitem[Zhou et~al.(2025)Zhou, Jia, Liu, and Fan]{zhou2025fuzz4cuda}
Yuhao Zhou, Peng Jia, Jiayong Liu, and Ximing Fan.
\newblock Fuzz4cuda: Fuzzing your nvidia gpu libraries through debug interface.
\newblock \emph{Computers \& Security}, page 104754, 2025.

\bibitem[Zhu et~al.(2026)Zhu, Chen, Fan, Ren, Wu, Chai, Rungrueangwutthinon,
  Ma, and Zou]{zhu2026cudabench}
Jiace Zhu, Wentao Chen, Qi~Fan, Zhixing Ren, Junying Wu, Xing~Zhe Chai,
  Chotiwit Rungrueangwutthinon, Yehan Ma, and An~Zou.
\newblock Cudabench: Benchmarking llms for text-to-cuda generation.
\newblock \emph{arXiv preprint arXiv:2603.02236}, 2026.

\end{thebibliography}
